\documentclass[10pt,journal,compsoc]{IEEEtran}
\ifCLASSOPTIONcompsoc
  \usepackage[nocompress]{cite}
\else
  \usepackage{cite}
\fi
\ifCLASSINFOpdf
\else
\fi
\usepackage{amsmath}
\usepackage{amssymb}
\usepackage{fixltx2e}
\usepackage{url}

\usepackage[normalem]{ulem}

\newcommand\mr[1]{{#1}}
\newcommand\shave[1]{{}}  

\usepackage[square,numbers]{natbib}
\usepackage{makecell} 

\usepackage[breaklinks=true,bookmarks=false,	
    colorlinks=true,
	linkcolor=blue,
	citecolor=blue,
	filecolor=blue]{hyperref}

\usepackage{booktabs}
\usepackage{adjustbox}
\usepackage{enumitem}
\usepackage[dvipsnames]{xcolor}
\usepackage{pifont}
\newcommand{\cmark}{\ding{51}}%
\newcommand{\xmark}{\ding{55}}%

\newcommand{\yhat}{\widehat{y}}
\renewcommand{\L}{\mathcal{L}}
\newcommand{\R}{\mathbb{R}}

\begin{document}
%
\title{A Survey of Self-Supervised and\\ Few-Shot Object Detection}
%
%
%
%

\author{Gabriel Huang, Issam Laradji, David Vázquez, Simon Lacoste-Julien, Pau Rodríguez
\IEEEcompsocitemizethanks{\IEEEcompsocthanksitem G. Huang is a PhD student at Mila \&  DIRO, Universit\'e de Montr\'eal, and a Visiting Researcher at ServiceNow. 
E-mail: gabriel.huang@umontreal.ca
\IEEEcompsocthanksitem I. Laradji, D. Vázquez and P. Rodríguez are with ServiceNow.
\IEEEcompsocthanksitem S. Lacoste-Julien is an Associate Professor at Mila \& DIRO, Universit\'e de Montr\'eal and holds a Canada CIFAR AI Chair.
}
\thanks{Manuscript received October 27, 2021; revised April 25, 2022.}
}

%
%


\markboth{IEEE Transactions on Pattern Analysis and Machine Intelligence}
{Huang \MakeLowercase{\textit{et al.}}: A Survey on Few-Shot and Self-Supervised Object Detection}
%



\IEEEtitleabstractindextext{%
\begin{abstract}
Labeling data is often expensive and time-consuming, especially for tasks such as object detection and instance segmentation, which require dense labeling of the image. While few-shot object detection is about training a model on \textit{novel} (unseen) object classes with little data, it still requires prior training on many labeled examples of \textit{base} (seen) classes. On the other hand, self-supervised methods aim at learning representations from unlabeled data which transfer well to downstream tasks such as object detection. Combining few-shot and self-supervised object detection is a promising research direction. In this survey, we review and characterize the most recent approaches on few-shot and self-supervised object detection. Then, we give our main takeaways and discuss future research directions. Project page:~\url{https://gabrielhuang.github.io/fsod-survey/}
\end{abstract}

\begin{IEEEkeywords}
self-supervised, few-shot, low-data, object detection, instance segmentation
\end{IEEEkeywords}}

\maketitle

\IEEEdisplaynontitleabstractindextext

%
\IEEEpeerreviewmaketitle

\IEEEraisesectionheading{\section{Introduction}\label{sec:introduction}}

\IEEEPARstart{T}{raditional} object detectors rely on large supervised object detection datasets such as PASCAL VOC~\cite{pascal} and MS COCO~\cite{coco}, which have over hundreds and thousands of annotated examples per object category. 
However, labeling data is often expensive and time-consuming. This is especially true in the case of object detection and instance segmentation, which require dense labeling of bounding boxes/masks for each object, a process that is slower and requires more annotator training than for object classification. 
Moreover, for fine-grained object detection applications such as plant or animal species recognition, pre-labeled datasets may not exist, and labels may have to be collected on the spot by expert annotators.

To try to solve these problems, few-shot object detection (FSOD) methods attempt to recognize \textit{novel} (unseen) object classes based only on a few examples, after training on many labeled examples of \textit{base} (seen) classes.
Until recently, the standard approach in few-shot object detection was to pretrain a backbone for ImageNet classification, then train an object detector on top of this backbone on the base classes, and finally finetune on the novel classes~\citep{metayolo,metarcnn,tfa,mpsr,metadetr}
However, because of the tremendous progress in learning self-supervised representations, several (few-shot) detection methods now initialize their backbone from representations pretrained with unsupervised pretext tasks on ImageNet and MS COCO~\citep{detreg,soco,insloc,resim,esvit,pinheiro2020unsupervised}.

The problem with typical self-supervised pretrained methods such as SimCLR~\citep{simclr} or MoCo~\citep{moco} is that they are geared towards classification, and often engineered to maximize \textit{Top-1} performance on ImageNet~\cite{densecl}. However, 
some of the learned invariances in classification (e.g. to translation) might not be desirable in localization tasks, and thus the representation might discard critical information for object detection. Moreover, it has been shown that higher ImageNet Top-1 accuracy does not necessarily guarantee higher object detection performance~\cite{densecl}.

In response to such shortcomings, there has been a growing number of methods for self-supervised object detection. These methods~\citep{updetr,densecl,insloc,resim,soco} not only attempt to remedy the shortcomings of classification-geared representations, but also pretrain more components in addition to the feature extractor, such as the region proposal network (RPN) and detection head, in the case of Faster R-CNN based methods.
In particular, the current state-of-the-art for FSOD on MS COCO is a method which does self-supervised pretraining of both the backbone and the object detector~\citep{detreg}.

\begin{figure*}
    \centering
    \includegraphics[width=\textwidth]{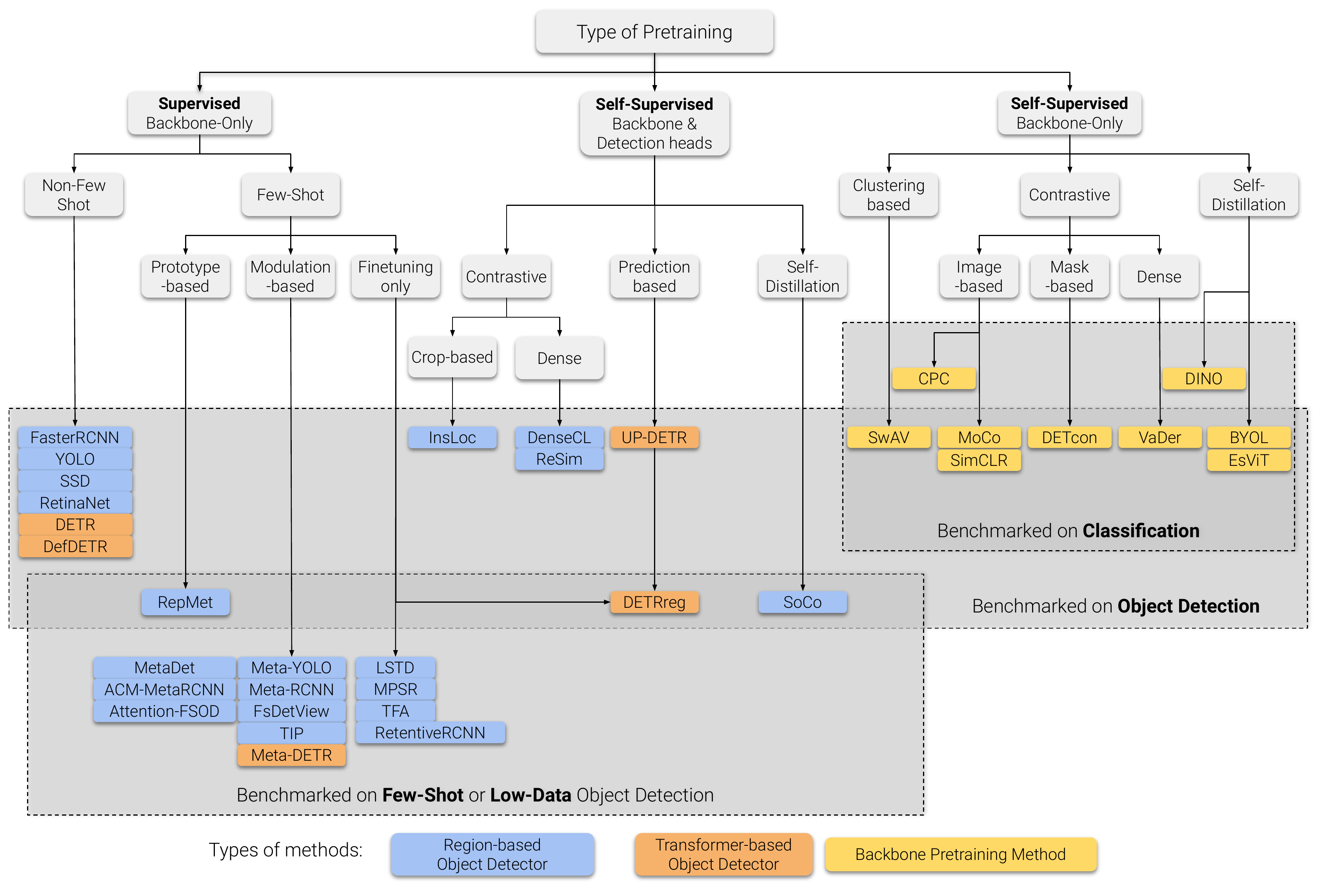}
    \caption{\textbf{A taxonomy of object detection methods reviewed in this survey.} We categorize them based on the following hierarchy: methods using supervised backbone pretraining, methods using self-supervised pretraining of backbone and detection heads, and self-supervised backbone pretraining methods. In parallel, we also tag (shaded rectangles) those methods depending on whether they have been benchmarked on regular object detection, few-shot/low-shot object detection, and ImageNet classification. As discussed in Section~\ref{sec:ss}, many self-supervised classification methods have also been used to initialize object detection backbones and evaluated on object detection benchmarks. DETReg~\citep{detreg}, which is a self-supervised object detection method, obtained state-of-the-art FSOD results on MS COCO and uses self-supervised pretraining of the entire architecture.}
    \label{fig:taxonomy}
\end{figure*}

Thus, this motivates a survey combining the most recent approaches on few-shot and self-supervised object detection, both of which having not been surveyed before (see Section~\ref{sec:related-surveys}). In the following sections, we briefly summarize key object detection concepts (Section~\ref{sec:background}). Then we review the few-shot object detection task and benchmarks (Section~\ref{sec:fsod}) and we discuss the most recent developments in few-shot object detection (Section~\ref{sec:fsod}) and self-supervised object detection pretraining (Section~\ref{sec:ss}). We conclude this survey by summing up the main takeaways, future trends, and related tasks (Sections~\ref{sec:takeaways}~and~\ref{sec:related-approaches}). We provide a taxonomy of popular few-shot and self-supervised object detection methods in Figure~\ref{fig:taxonomy}, on the base of which this survey is structured.

\section{Related Surveys\label{sec:related-surveys}}
\citet{jiao2019survey}~and~\citet{zaidi2021survey} survey modern deep-learning based object detection methods.
They review several state-of-the-art backbones and compare their parameter counts. They present common datasets benchmarks and evaluation metrics. \citet{jiao2019survey} categorize object detectors into single-stage and two-stage detectors, and report their respective performances in a large table. \citet{zaidi2021survey} is more recent and discusses backbone choices extensively, including transformer-based backbones such as Swin transformers~\citep{swin}, and lightweight architectures such as MobileNet~\citep{howard2017mobilenets} and SqueezeNet~\citep{iandola2016squeezenet}. Although both works focus on traditional object detection, they briefly mention weakly-supervised, few-shot and unsupervised object detection as future trends. Note that these surveys do not cover the newer transformer-based object detectors such as DETR~\citep{detr} or Deformable DETR~\citep{deformabledetr}, which we will briefly introduce in this survey.

\citet{jing2020self} present a survey on self-supervised visual feature learning. They perform an extensive review of self-supervised pretext tasks, backbones, and downstream tasks for image and video recognition. In this work, we also introduce a number of generic self-supervised pretraining techniques but we focus on methods particularly designed for object detection.

Regarding few-shot classification, a simpler task than few-shot object detection, \citet{chen2018closer} introduce a comparative analysis of several representative few-shot classification algorithms. \citet{wang2020generalizing} propose a more extensive survey on methods and datasets. However, they do not explore few-shot object detection methods.

\citet{khan2021transformers} show that transformers have achieved impressive results in image classification, object detection, action recognition and segmentation.
In this survey, we review object detection methods using transformers as backbones and as detection heads with DETR and variants~\citep{detr,deformabledetr}. We also discuss the emergent properties of visual transformers as showcased by~\citet{dino}.


\section{Background on Object Detection\label{sec:background}}

\subsection{Key Concepts}
\begin{figure*}
    \centering
    \includegraphics[width=0.9\linewidth]{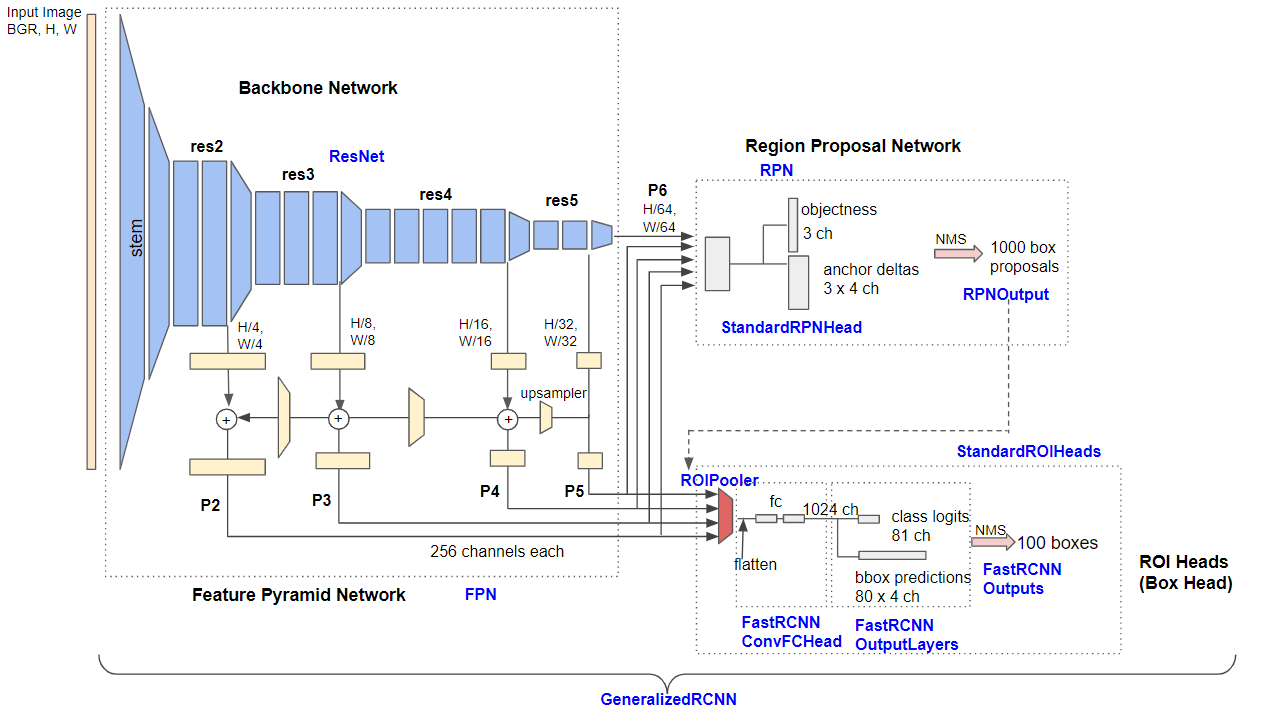}
    \caption{\textbf{A Faster R-CNN with Feature Pyramid Network.} The input image is fed to the backbone network, then the feature pyramid network (light yellow) computes multi-scale features. The region proposal network proposes candidate boxes, which are filtered with non-maximum suppression (NMS). Features for the remaining boxes are pooled with RoIAlign and fed to the box head, which predicts object category and refined box coordinates. Finally, redundant and low-quality predictions are removed with NMS.
    Blue labels are class names in the detectron2 implementation. Figure courtesy of Hiroto Honda. \protect\url{https://medium.com/@hirotoschwert/digging-into-detectron-2-47b2e794fabd}}
    \label{fig:fasterrcnndetectron2}
\end{figure*}

\begin{figure*}
    \centering
    \includegraphics[width=0.8\linewidth]{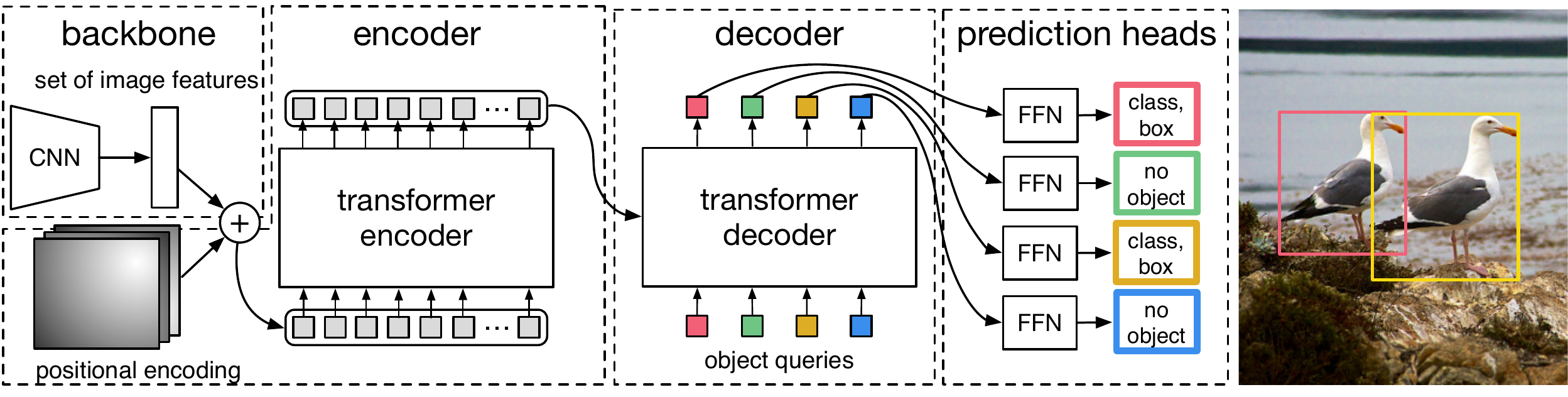}
    \caption{\textbf{The DETR object detector}. The image is fed to the backbone, then positional encodings are added to the features and fed to the transformer encoder. The decoder takes as input object query embeddings, \mr{cross-attends to the encoded representation while performing self-attention on the transformed query embeddings}, and outputs a fixed number of object detections, which are finally thresholded, without need for NMS~\citep{detr}. Image courtesy of~\citet{detr}.}
    \label{fig:detr}
\end{figure*}

For clarity, we start by reviewing key concepts in object detection, and introduce relevant vocabulary. 
Readers already familiar with object detection can skip directly to Sections~\ref{sec:fsod}~and~\ref{sec:ss} for few-shot and self-supervised object detection.
We illustrate those concepts in the context of Faster R-CNN~\citep{fasterrcnn} with Feature Pyramid Network~\citep{fpn}, a multi-scale two-stage object detector represented in Figure~\ref{fig:fasterrcnndetectron2}, and DETR~\citep{detr} represented in Figure~\ref{fig:detr}. A more in-depth analysis of object detection concepts can be found in the object detection surveys by~\citet{jiao2019survey,zaidi2021survey}.

\textbf{Object detection} is the task of jointly localizing and recognizing objects of interest in an image. Specifically, the object detector has to predict a bounding box around each object, and predict the correct object category. Object detectors are traditionally trained on labeled object detection datasets such as PASCAL VOC~\citep{pascal} or MS COCO~\citep{coco}; the objects of interest are simply the categories that the model is trained to recognize.

The \textbf{backbone} network is a feature extractor which takes as input an RGB image and outputs one or several feature maps~\citep{fpn}. 
Typically, the backbone is a residual network such as the ResNet-50~\citep{he2016deep}, and is pretrained on ImageNet classification before finetuning it to downstream tasks~\citep{tfa,metarcnn,detreg}. 
Alternatively, an increasing number of works have considered using visual transformers instead~\citep{yolov3,esvit,dino}.
The RGB image is a 3D tensor in $\R^{W\times H\times 3}$, where typically $W=H=224$ for classification, and $W,H\approx 1300,800$ for object detection (as per Detectron2's\footnote{\url{https://github.com/facebookresearch/detectron2}} default parameters). For few-shot object detection, a ground-truth mask delimiting the support object is sometimes appended to a fourth channel of the image, and the backbone is modified to take as input tensors in $\R^{W\times H\times 4}$.

\textbf{Single-scale} features consist of a single 3D tensor obtained by taking the outputs of a certain backbone layer. Typically, the C4 layer (corresponding to the output of ``res5'' the 4th residual block) of the ResNet-50 is used for object detection. The feature map is of size $Z\in\R^{w\times h\times c}$ where $c$ is the number of channels and $w,h$ are the spatial dimensions of the feature map, which are much smaller than the image due to strided convolutions. 

\textbf{Multi-scale} features consist of several 3D tensors at different scales. Merely combining several layer outputs from the backbone would result in the high-resolution lower layers having limited semantic information. A common solution is to implement top-down and lateral pathways using a Feature Pyramid Network (\textbf{FPN})~\citep{fpn} to propagate information from the higher-levels of the backbone back to the lower levels (illustrated in Figure~\ref{fig:fasterrcnndetectron2}).

\textbf{Faster R-CNN}~\citep{fasterrcnn}, represented in Figure~\ref{fig:fasterrcnndetectron2}, is a popular two-stage object detector. To detect objects, they start by feeding an image to the backbone to get single or multi-scale features. Then, they apply the following two stages:
\begin{itemize}
    \item \textit{Stage 1}: They feed the features to the \textbf{Region Proposal Network} (RPN) to extract \textbf{object proposals}, which are bounding boxes susceptible to contain objects. The object proposals are predicted at predefined locations, scales and aspect ratios (known as \textbf{anchors}), refined using a regression head (anchor deltas), and scored for ``\textbf{objectness}''. They use Non-Maximum Suppression (\textbf{NMS}) to remove redundant and low-quality object proposals.
    \item \textit{Stage 2}: For each object proposal they extract a pooled feature map by resampling the features inside its bounding box to a fixed size, using RoIAlign or ROIPool (pooling strategies). For multiscale-features, the appropriate level is picked using a heuristic.
    Then, they feed the pooled features into the \textbf{Box Head} or \textbf{Region-of-Interest} (ROI) head, which predicts the object category and refines the bounding box with another regression head. Finally, they run NMS again to remove redundant and low-confidence predictions. 
\end{itemize}
In this survey, we will refer to the union of the RPN and box head as the \textbf{detection heads}.

\textbf{Mask R-CNN}~\citep{maskrcnn} is an improvement on top of Fast R-CNN to solve instance segmentation. At the simplest level, Mask R-CNN predicts segmentation masks for each detected instance, additionally to the bounding box and class predictions.

\textbf{Single-stage} object detectors, such as You Only Look Once (\textbf{YOLO})~\citep{yolo,yolov3}, Single-Shot Detector (\textbf{SSD})~\citep{ssd}, and newer methods such as RetinaNet~\citep{retinanet} and CenterNet~\citep{centernet}, are generally simpler and faster than two-stage detectors, at the cost of lower prediction quality. Single-stage detectors directly predict objects at predefined locations from the feature map, with the possibility of subsequently refining the box locations and aspect ratios. Please refer to object detection surveys for an in-depth review~\citep{jiao2019survey,zaidi2021survey}.

\textbf{DETR}~\citep{detr} represented in Figure~\ref{fig:detr}, which stands for DEtection TRansformer, is a recent transformer-based architecture for end-to-end object detection.
Notably, DETR has a much simpler overall architecture than Faster R-CNN and removes the need for the NMS heuristic --which is non-differentiable-- by learning to remove redundant detections. However, being set-based, DETR relies on the Hungarian algorithm~\citep{munkres1957algorithms} for computing the prediction loss, which has been shown to be difficult to optimize~\citep{sun2020rethinking}.

To detect objects, they start by feeding an image to the backbone to get features. Then, they feed the features to the transformer \textbf{encoder} to obtain encoded features. 
Finally, they feed 100 (or any number of) learned ``query embeddings'' to the transformer \textbf{decoder}. The transformer decoder attends to the encoded features and outputs up to 100 predictions, which consist of bounding box locations, object categories and confidence scores. The highest-confidence predictions are returned. There is no need for removing redundant predictions using NMS, as the model learns to jointly make non-redundant predictions thanks to the Hungarian loss. 
During training, the \textit{Hungarian loss} is computed by finding the optimal matching between detected and ground-truth boxes in terms of box location and predicted class. The loss is minimized using \mr{stochastic} gradient descent (SGD).
We will also refer to the union of transformer encoder and decoder as the \textbf{detection heads}.

\textbf{Deformable DETR}~\citep{deformabledetr} is a commonly used improvement over DETR. Deformable DETR uses multi-scale deformable attention modules, which can attend to a small set of learned locations over multiple feature scales, instead of attending uniformly over a whole single-scale feature map. The authors manage to train their model using 10 times \mr{fewer} epochs than DETR~\citep{deformabledetr}.

\subsection{Datasets and Evaluation Metrics}
The most popular datasets for traditional object detection are PASCAL VOC~\citep{pascal} and MS COCO~\citep{coco}. Since they have already been widely discussed in the literature, we refer the reader to previous object detection surveys~\citep{jiao2019survey,zaidi2021survey}.
PASCAL VOC and MS COCO have also been adopted by the few-shot object detection (FSOD) and self-supervised object detection (SSOD) communities. We provide an extensive discussion on their use as few-shot benchmarks in Section~\ref{sec:fsod-datasets}.
Please also refer to Section~\ref{sec:fsod-datasets} for a detailed explanation of the mean average precision (mAP) evaluation metric and the differences between PASCAL VOC and MS COCO implementations.

\section{Few-Shot Object Detection\label{sec:fsod}}

Informally, few-shot object detection (FSOD) is the task of learning to detect new categories of objects using only \textit{one} or a \textit{few} training examples per class. In this section, we describe the FSOD framework, its differences with few-shot classification, common datasets, evaluation metrics, and FSOD methods. We provide a taxonomy of popular few-shot and self-supervised object detection methods in Figure~\ref{fig:taxonomy}.

\subsection{FSOD Framework\label{sec:fsodframework}}

\begin{figure}
    \centering
    \includegraphics[width=\linewidth]{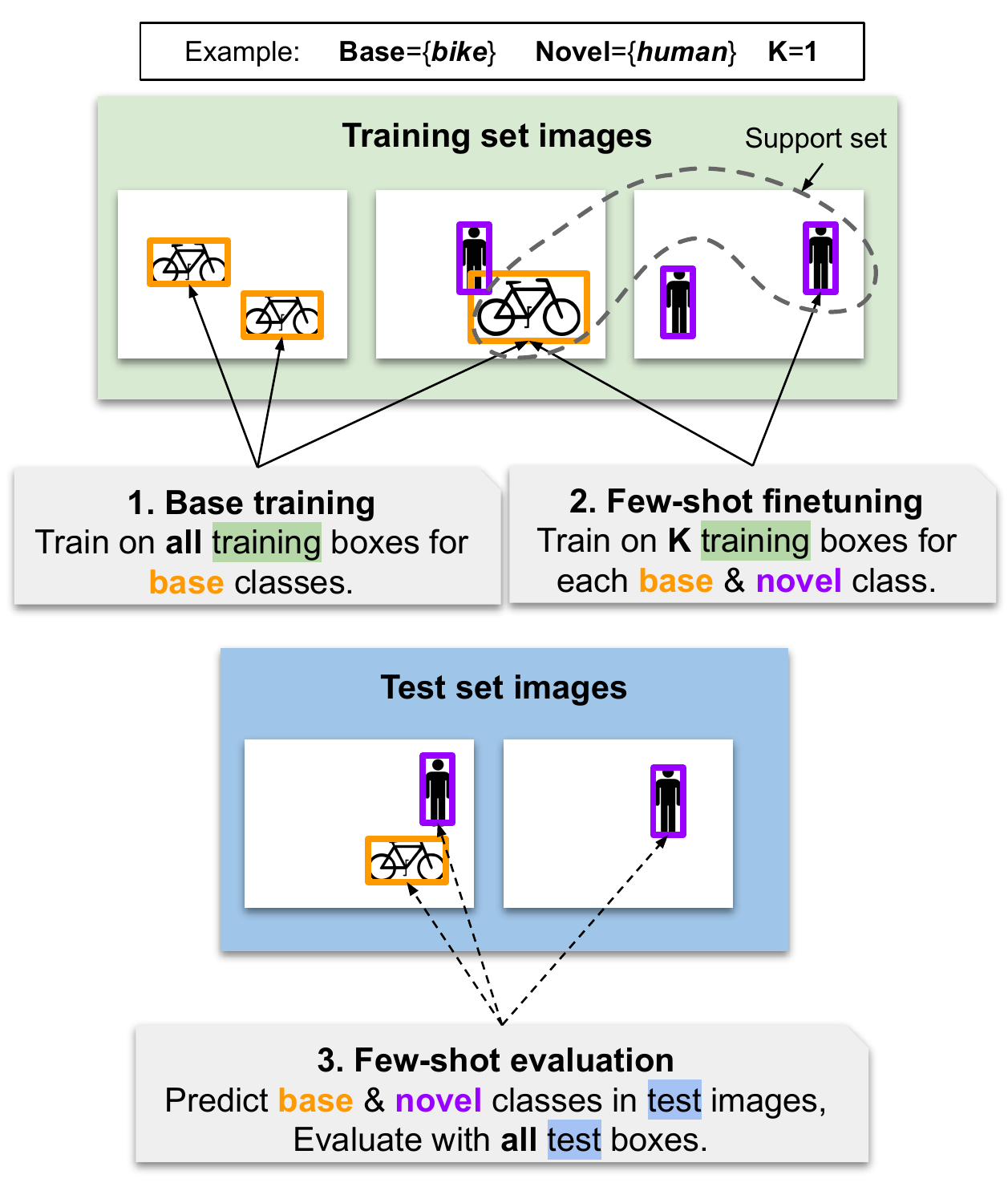}
    \caption{\textbf{Few-shot object detection protocol}, as proposed by~\citet{metayolo}. During base-training, the method is trained on base classes. Then during few-shot finetuning, the model is finetuned or conditioned on the support set. Finally, during few-shot evaluation, the method is evaluated on base and novel class detection.}
    \label{fig:fsod}
\end{figure}

We formally introduce the dominant FSOD framework, as formalized by~\citet{metayolo} (\textbf{Figure~\ref{fig:fsod}}). FSOD partitions objects into two disjoint sets of categories: \textbf{base} or known/source classes, which are object categories for which we have access to a large number of training examples; and \textbf{novel} or unseen/target classes, for which we have only a few training examples (shots) per class. 
%
In the vast majority of the FSOD literature, we assume that the object detector's backbone has already been pretrained on an image classification dataset such as ImageNet (usually a ResNet-50 or 101). 
Then, the FSOD task is formalized as follows:
\begin{enumerate}[label={(\arabic*)}]
    \item \textbf{Base training.}\footnote{In the context of self-supervised learning, base-training may also be referred to as \textit{finetuning} or \textit{training}. This should not be confused with \textit{base training} in the meta-learning framework; rather this is similar to the meta-training phase~\citep{finn2017model}.} 
    Annotations are given only for the base classes, with a large number of training examples per class (\textit{bikes} in the example).
    We train the FSOD method on the base classes.
    \item \textbf{Few-shot finetuning.} 
    Annotations are given for the \textit{support set}, a very small number of training examples from \textit{both} the base and novel classes (one \textit{bike} and one \textit{human} in the example).
    Most methods finetune the FSOD model on the support set, but some methods might only use the support set for conditioning during evaluation (finetuning-free methods).
    \item \textbf{Few-shot evaluation.} 
    We evaluate the FSOD to jointly detect base and novel classes from the test set (few-shot refers to the size of the support set). The performance metrics are reported separately for base and novel classes.
    Common evaluation metrics are variants of the mean average precision: mAP50 for Pascal and COCO-style mAP for COCO. They are often denoted bAP50, bAP75, bAP (resp. nAP50, nAP75, nAP) for the base and novel classes respectively, where the number is the IoU-threshold in percentage (see Section~\ref{sec:fsod-eval} for full explanation).
\end{enumerate}

\mr{We encourage researchers to report both base \textit{and} novel class performance, a setting sometimes called Generalized FSOD~\citep{retentivercnn}}. Note that ``training'' and ``test'' set refer to the splits used in traditional object detection. Base and novel classes are typically present in both the training and testing sets; however, the novel class annotations are filtered out from the training set during base training; during few-shot finetuning, the support set is typically taken to be a (fixed) subset of the training set; during few-shot evaluation, all of the test set is used to reduce uncertainty~\citep{metayolo}.

\paragraph*{\textbf{Special case—no fine-tuning}} Conditioning-based methods skip the fine-tuning step; instead, novel examples are used as support examples to condition the model, and predictions are made directly on the test set. In practice, the majority of conditioning-based methods reviewed in this survey do benefit from some form of finetuning.

\mr{\paragraph*{\textbf{Special case—no base classes}} The standard FSOD framework might seem counter-intuitive as it assumes a large set of annotated base class images. DETReg~\citep{detreg} have investigated  removing the base-training phase and replacing it with self-supervised pretraining, relying solely on a small number of novel labels, which came at the cost of lower nAP.}

\subsection{Important Differences with Few-Shot Classification.} \label{sec:fscvsfsod}

\begin{figure}[!t]
    \centering
    \includegraphics[width=\linewidth]{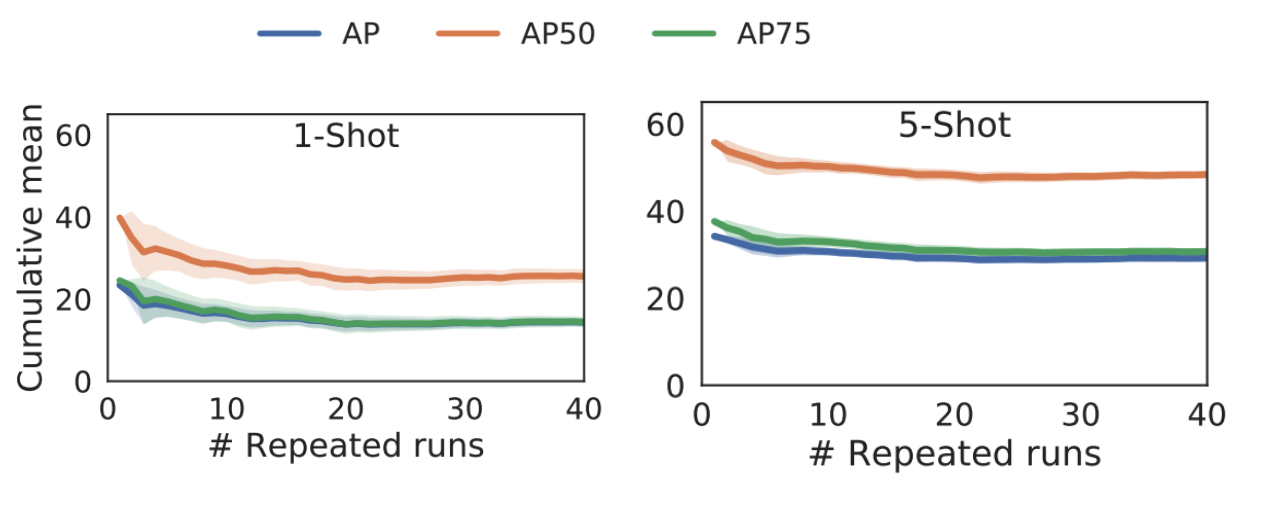}
    \caption{\textbf{Importance of evaluating over several episodes.} The nAP, nAP50 and nAP75 of PASCAL VOC Split-1 are averaged using a variable number of episodes. Note how the means and variances only become stable after around 20 episodes. Figure courtesy of~\citep{tfa}.}
    \label{fig:pascalnoise}
\end{figure}

As this may not be obvious to readers unfamiliar with both fields, we explicit several practical differences between the FSOD finetuning-based paradigm and the learning-to-learn paradigm~\citep{ravi2016optimization,matchingnet,chen2018closer,huang2019few} commonly used in few-shot classification (FSC):
\begin{itemize}
    \item \textbf{Several objects per image.} In FSOD there can be several instances from base and novel classes in the same image, whereas FSC assumes only one dominant object per image. While FSC filters out all novel \textit{images} during base-training, FSOD removes only the novel object annotations but keeps the images that also contain base objects.
    \item \textbf{Joint prediction on base and novel.} During few-shot finetuning and evaluation, both base and novel classes are present and have to be jointly detected.
    On the contrary, few-shot classifiers are typically only finetuned and evaluated on novel classes. 
    Note however that many papers only report average precision for novel classes under metric names such as \textit{nAP} or \textit{nAP50}.
    \item \textbf{Learning-to-learn vs. finetuning.}
    Gradient-based few-shot classification methods such as MAML~\citep{finn2017model} or Learning-to-Optimize~\citep{ravi2016optimization} rely heavily on the \textit{learning-to-learn} paradigm; during (meta)training, N-way K-shot episodes are generated by sampling a small training set (support set) and a small validation set (query set) from the base classes. The classifier is finetuned with gradient descent on the support set, makes predictions on the query set, and the query-set loss is minimized using gradient descent, which propagates the gradient through support set tuning. 
    %
    On the contrary, a majority of FSOD methods do not generally consider episodes or backpropagate through gradient descent. Pure finetuning FSOD methods~\cite{tfa,lstd,mpsr,detreg} are first trained on all base classes during base training, and finetuned only once on a fixed support set before few-shot evaluation.
    Moreover, because the few-shot finetuning step (e.g., optimizer learning rates) and the pre-finetuning weights are not calibrated over several episodes using learning-to-learn on a separate query set, they might not be optimal. This is partially mitigated by hyperparameter tuning, which can help find optimal learning rates, but not find the optimal pre-finetuning weights.
    \item \textbf{Episodic Evaluation.} For FSC evaluation, several episodes are sampled from the novel classes~\citep{matchingnet,oreshkin2018tadam,rodriguez2020embedding}; the classifier is finetuned on the support set and classifies the query set, and the results are averaged over hundreds of runs, which have the advantage of reducing variance and estimating confidence intervals~\citep{chen2018closer,huang2019few}. On the contrary, each of Kang's splits~\citep{metarcnn} feature only \textit{one} fixed support set (the exact instances are prespecified), which is known to cause overfitting and overestimating performance especially in the case of 1-shot object detection, when the support set is tiny~\citep{tfa}. See Figure~\ref{fig:pascalnoise} for the impact of using several episodes on PASCAL VOC.
    In response to those issues, \citet{tfa} \mr{sample multiple support sets (30 seeds) to lower the variance of the benchmark.}
    \item \textbf{Separate validation and test sets.} Whereas FSC methods generally validate and test on separate splits for common benchmarks such as Omniglot~\citep{omniglot}, \textit{mini}ImageNet~\citep{matchingnet}, or Synbols~\cite{Lacoste2020SynbolsPL}, FSOD detection methods follow the standard practice in object detection of training on the union of the training and validation sets, and using the test set for both hyperparameter tuning and evaluation, which inevitably leads to overestimating the generalization ability of the methods.
\end{itemize}

\begin{table*}[!t]
    \centering
    \caption{\textbf{Common FSOD benchmarks}. 
    %
    Image counts are after filtering out the ones containing no relevant bounding boxes.
    }

    \label{tab:fsd-benchmarks}

    \begin{tabular}{lrrrrlrrrrr}
\toprule
Benchmark 
    & \multicolumn{2}{c}{Classes}
    & \multicolumn{2}{c}{1. Base Training}
    & \multicolumn{3}{c}{2. Few-shot Finetuning}
    & \multicolumn{3}{c}{3. Few-shot Evaluation} \\
    \cmidrule(r){2-3}  \cmidrule(lr){4-5}  \cmidrule(lr){6-8}  \cmidrule(l){9-11}
    & Base & Novel & \#images & \#\textbf{base}-bb
    & \#shots & \#\textbf{base}-bb & \#\textbf{novel}-bb
    & \#images & \#\textbf{base}-bb & \#\textbf{novel}-bb\\
    \midrule
PASCAL VOC/split 1 & 15 & 5 & 14,631 & 41,084 &  1:2:3:5:10  & 15--150 & 5--50 & 4,952 & 13,052 & 1,924 \\
PASCAL VOC/split 2 & 15 & 5 & 14,779 & 40,397 &  1:2:3:5:10  & 15--150 & 5--50 & 4,952 & 12,888 & 2,088 \\
PASCAL VOC/split 3 & 15 & 5 & 14,318 & 40,511 & 1:2:3:5:10  & 15--150 & 5--50 & 4,952 & 13,137 & 1,839 \\
MS COCO 2014 & 60 & 20 & 98,459 & 367,702 & 10:30 & 600-1,800 & 200-600 & 5,000 & 15,318 & 20,193 \\
LVIS v0.5 & 776 & 454 & 68,568 & 688,029 & 8.57 (variable) & 7,760 & 2,786 & 5,000 & 50,334 & 429 \\ 
\bottomrule
    \end{tabular}

\end{table*}

\subsection{FSOD Datasets\label{sec:fsod-datasets}}

We describe the dominant FSOD benchmarks, as introduced by Meta-YOLO~\citep{metayolo} and improved by TFA (Two-stage Fine-tuning Approach)~\citep{tfa} to mitigate variance issues.
These are also the benchmarks that we will use to compare FSOD methods in Table~\ref{tab:fsod}. 
We compute data statistics in \textbf{Table~\ref{tab:fsd-benchmarks}}.\footnote{Note that the reported number of images is after removing those containing no relevant annotations (e.g., for base training, the images which contained only novel objects are removed).} Caveats and future best practices are discussed in Section~\ref{sec:fsod-dataset-discussion} and Section~\ref{sec:takeaways-eval}.

\subsubsection{PASCAL VOC~\citep{pascal}} PASCAL VOC is arguably one of the most popular smaller benchmarks for traditional object detection. 
For few-shot object detection, the object categories are split between \textbf{15 base classes} and \textbf{5 novel classes}.
Three different base/novel splits are usually considered, denoted splits 1, 2 and 3.
For base training, the training and validation (\textit{trainval}) images from VOC2007 and VOC2012 are used, which might lead to overfitting (Section~\ref{sec:takeaways-eval}). 
\shave{
This amounts to \textbf{40k base boxes} spread over 15k images for an average of  \textbf{2700 boxes / base class}, with 2.8 boxes / image.}
For few-shot finetuning, a fixed subset of the VOC2007 and VOC2012 \textit{trainval} sets is taken as the support set.
\citet{metayolo} consider the \textbf{1,2,3,5,10}-shot settings, which correspond to 15-150 base bounding boxes and 5-50 novel bounding boxes. The fact that the instances are fixed may lead to overestimating performance (Section~\ref{sec:fsod-dataset-discussion}).
For few-shot evaluation, roughly 5k images from the VOC2007 test set are considered.
\shave{, with 13k base and 2k novel boxes, which adds up to an average of \textbf{870 boxes/base} and \textbf{400 boxes/novel} class. The method has to detect both base and novel classes, and several evaluation metrics are reported separately for base and novel classes: mAP50, mAP75, and COCO-style mAP.}
The main comparison metric is the novel mAP50, with  mAP75 and COCO-style mAP often reportted.

\subsubsection{MS COCO~\citep{coco}} MS COCO or Microsoft Common Objects in COntext~\citep{coco}, is another popular benchmark for object detection.
For FSOD, the object categories are split between \textbf{20 novel classes} which are shared with PASCAL VOC, and the remaining \textbf{60 base classes}. Following \citet{metayolo} and \citet{tfa}, a 5k subset of the COCO2014 validation images are used for few-shot evaluation (denoted \textit{val5k}), while the remaining COCO2014 training and validation images are used for base-training and few-shot finetuning (denoted \textit{trainvalno5k}).
\shave{
Specifically for base training, roughly \textbf{367k base boxes} are considered (10 times more than PASCAL VOC), spread over 98k images from \textit{trainvalno5k}. This is an average of \textbf{6k boxes / base class} and 3.7 boxes/image.
For few-shot finetuning, a fixed subset of \textit{trainvalno5k} is taken as the support set.
\citet{metayolo} consider the \textbf{10,30}-shot settings, which correspond to 600-1800 base boxes and 200-600 novel boxes, and suffer less from overfitting than PASCAL VOC~\citep{tfa} despite also using fixed instances.
For few-shot evaluation, the \textit{val5k} are used, consisting in 15k base boxes and 20k novel boxes, which amounts to \textbf{250 boxes / base} class and \textbf{1k boxes / novel} class. 
}
Several evaluation metrics are reported separately for base and novel categories: COCO-style mAP, mAP50, mAP75, and mAP for small, medium and large objects. The main comparison metric is novel mAP.

\subsubsection{LVIS~\citep{gupta2019lvis}} LVIS or Large Vocabulary Instance Segmentation~\citep{gupta2019lvis} is a newer object detection dataset featuring 1230 classes, categorized as \textit{frequent}, \textit{common} (10 or more annotations) and \textit{rare} (\mr{fewer} than 10).
TFA~\citep{tfa} have proposed using the \textit{v0.5} version of this dataset for FSOD, dividing it into \textbf{776 base classes} (frequent and common) and \textbf{454 novel classes}, making it by far the FSOD benchmark with the most number of categories (10 times more categories than COCO, 50 times more than Pascal).\footnote{For this description, we follow the reference implementation from TFA\footnote{\url{https://github.com/ucbdrive/few-shot-object-detection}} as we could not find all the details in the paper.}
\shave{
For base training, \textbf{688k base boxes} are considered, spread over 69k images from the training set, which amounts to \textbf{887 boxes / base class} and 10 boxes / image. 
For few-shot finetuning, up to 10 shots from the training set are considered, depending on the number of available examples. This corresponds to an average of 8.57 boxes/class, spread over 7.7k base boxes and 2.8k novel boxes. 
For few-shot evaluation, the validation set of LVIS is used, consisting of 50,334 base and 429 novel boxes, spread over 5,000 evaluations images. 
}
Evaluation metrics are COCO-style mAP, mAP50 and mAP75, reported separately for frequent, common, rare objects, and also aggregated over all three categories.

\subsubsection{Discussion \label{sec:fsod-dataset-discussion}}

We discuss some of the advantages and issues associated with the aforementioned benchmarks (see also Section~\ref{sec:takeaways-eval}).

\paragraph*{\textbf{Overfitting issues}} 
For PASCAL VOC, the support set is very small (20-200 examples) and the specific instances are predefined by Kang's splits. This can cause overfitting and result in overestimating the novel average precision, especially in the 1-shot case, an issue illustrated by TFA~\citep{tfa} in Figure~\ref{fig:pascalnoise}.
To mitigate this issue, TFA~\citep{tfa} propose to \mr{randomly sample multiple support sets (30 seeds)} and to average the results in a benchmark which we will denote the \textbf{TFA-splits} as opposed to the benchmark using a single fixed support set, which we will denote the \textbf{Kang-splits}.
This is a good first step, but not as reliable as the common practice in few-shot classification of averaging metrics over 600 episodes~\citep{chen2018closer}.

\paragraph*{\textbf{Reliability}} With a substantial support set of 800-2400 bounding boxes for few-shot finetuning, and plenty of few-shot evaluation boxes for base and novel categories, MS COCO is arguably the most reliable benchmark for comparing different methods. In fact, we sort methods according to 30-shot MS COCO nAP in Table~\ref{tab:fsod}. Because the 20 novel classes were chosen to in common with Pascal, a very natural benchmark to consider is the MS COCO$\longrightarrow$Pascal cross-domain scenario, which has been considered by some of the earlier and subsequent works~\citep{lstd,mpsr,fgn}.

\paragraph*{\textbf{Limitations of LVIS v0.5}} With 1230 classes, LVIS has more than an order of magnitude more object categories than MS COCO, and a lot of potential for few-shot object detection. 
However there are some shortcomings with directly using TFA's splits.
One problem is that only 705 out of 776 base classes and 125 out of 454 novel classes appear in the validation set, which means that the majority of novel classes will never be evaluated on\footnote{As found by running TFA's data loader: \url{https://github.com/ucbdrive/few-shot-object-detection}}
Moreover, because there are almost 100 more times base than novel boxes in the validation set, performance is completely dominated by base objects for metrics aggregated on base and novel classes. Finally, even the 125 novel classes that are evaluated on only have an average of 3.4 boxes / class, which means evaluation is potentially very noisy.
Due to those issues, we do not recommend the current TFA splits for evaluating few-shot object detection methods. However, LVIS—\mr{especially LVIS v1.0 which has more data and stricter quality control}—has a lot of potential in FSOD given the large diversity of objects. Therefore, proposing more balanced splits and evaluation sets would definitely be beneficial to the FSOD community.
\mr{\paragraph*{\textbf{Class overlap}}``Novel'' FSOD classes may overlap with ImageNet classes used for backbone pretraining, potentially leading to performance overestimation. For instance, all of the 20 PASCAL VOC categories—used as MS COCO novel classes—except ``person'' appear in some form inside ImageNet. Related categories may be parts of one another (car mirror, aircraft carrier), subcategories (sport car, rocking chair, flowerpot vs. potted plant), or functionally similar objects (sofa vs. studio couch, motorbike vs. scooter). Because of the large number of ImageNet labels, we acknowledge the challenge of designing novel classes with no overlap with ImageNet for general-purpose object detectors. However, for niche FSOD applications, we do encourage practitioners to sanitize backbone and FSOD datasets for cross-contamination.}
%

\subsection{FSOD Evaluation Metrics\label{sec:fsod-eval}}

By design, object detectors such as Faster R-CNN and DETR output a \textit{fixed} number of predictions paired with a confidence score. This means they use a threshold to cut off low-confidence predictions.
Therefore, they have to trade off between using a higher threshold, which will lead to higher precision (most predicted objects are actually real objects) but low recall (miss out many of the real objects); and using a lower threshold, which could increase recall at the expense of precision.
 
The \textbf{mean average precision (mAP)} is a standard metric for evaluating object detection and FSOD methods, and is defined as the mean of the individual average precisions (AP) for each object class. The individual APs are defined as the area under the precision-recall curve -- discussed below -- which can be plotted by varying the confidence threshold of the object detector.

To compute the AP for a class, we first rank the detections for this class by decreasing confidence. 
Starting from the top-ranked detections ($k=1$), we consider them as \textit{True Positives} if their intersection over union (IoU) with any ground-truth true object is above a given IoU-threshold (typically 50\% or 75\%). If the IoU is below the threshold or the ground-truth has already been detected, then we consider them to be \textit{False \mr{Positives}}. 
For each rank $k$, corresponding to a different choice of confidence-threshold, we can compute the precision@$k$, a measure of relevance defined as the number of true positives among the top-$k$ boxes divided by $k$, and the recall@$k$, a measure of sensitivity defined as the number of true positives among the top-$k$ boxes divided by the total number of ground truth boxes.

\begin{figure}
    \begin{center}
        \includegraphics[width=0.9\linewidth]{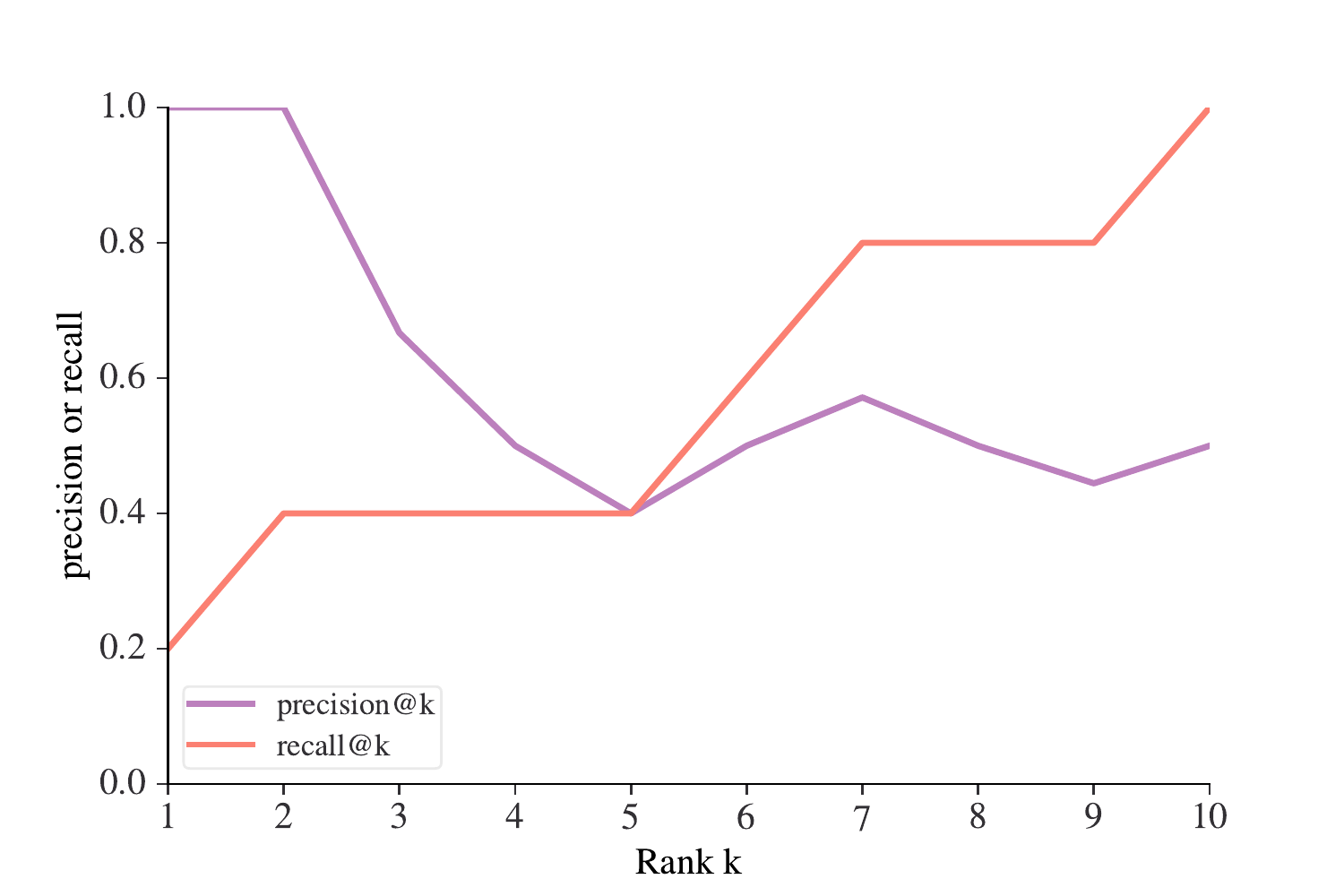}
        \includegraphics[width=0.9\linewidth]{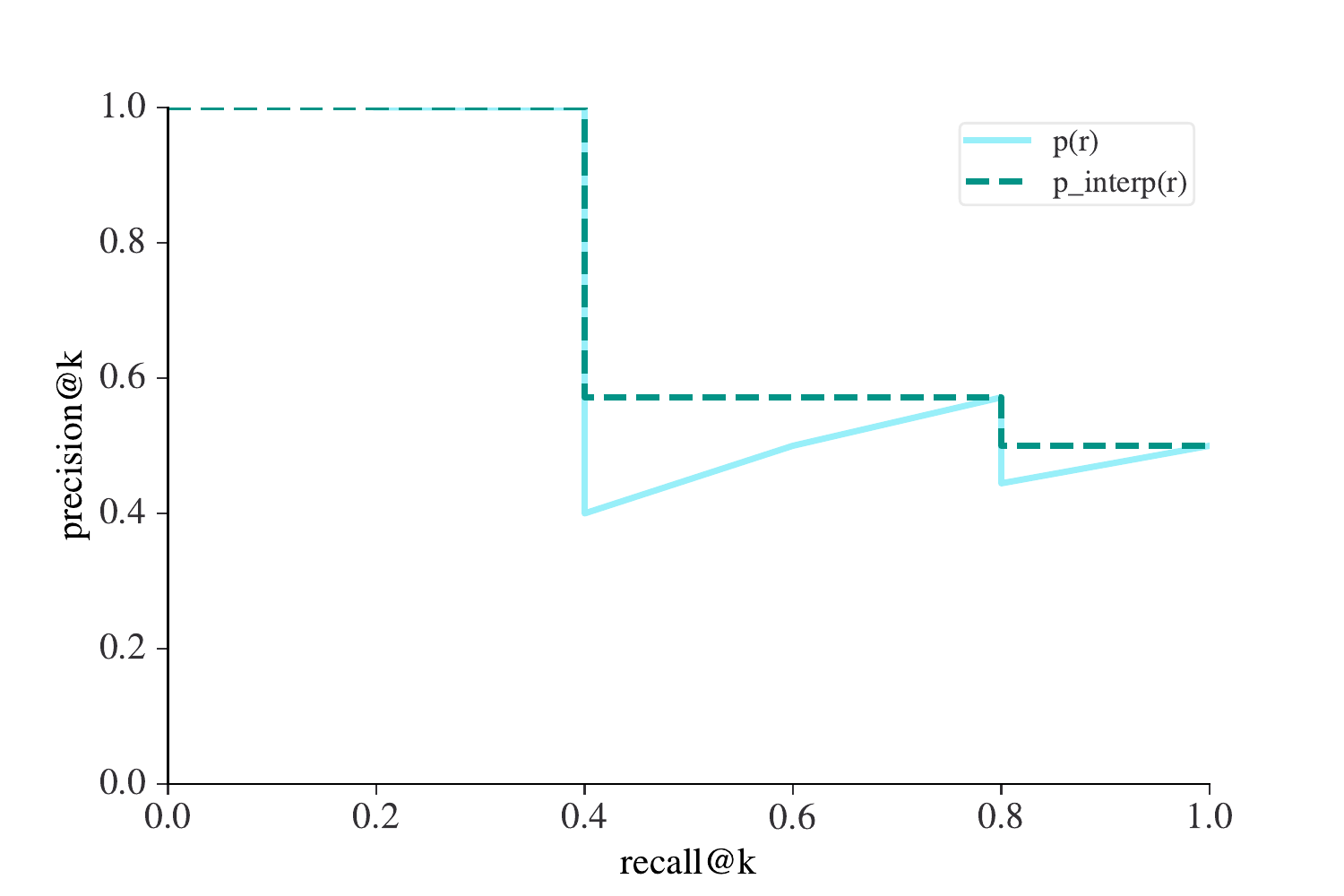}
        \caption{\textbf{Top}: precision@k and recall@k as a function of $k$ the number of boxes considered. \textbf{Bottom}: precision@k and interpolated-precision@k as functions of the recall@k (precision-recall curve)}
    \label{fig:precision-recall}
    \end{center}
\end{figure}

We give an example of AP computation in Table~\ref{table:precision-recall}.
\begin{table}
    \begin{center}
    \caption{\textbf{Example computation of precision@k and recall@k.} There are 10 detections ranked by decreasing confidence and 5 ground-truth boxes. Example~from \protect\url{https://jonathan-hui.medium.com/map-mean-average-precision-for-object-detection-45c121a31173}}
    \label{table:precision-recall}
    \begin{adjustbox}{width=0.8\linewidth}
    \begin{tabular}{cccc}
        \toprule
        Rank $k$ & True Positive? & Precision@$k$ & Recall@$k$ \\
        \midrule
        1   &  True & 1.0  & 0.2  \\
        2   &  True & 1.0  &  0.4 \\
        3   &  False & 0.67  & 0.4  \\
        4   &  False & 0.5  & 0.4  \\
        5   &  False & 0.4  & 0.4  \\
        6   &  True &  0.5 & 0.6  \\
        7   &  True &  0.57 & 0.8   \\
        8   & False  &  0.5 & 0.8  \\
        9   & False &  0.44 & 0.8  \\
        10   & True  & 0.5  & 1.0  \\
        \bottomrule
    \end{tabular}
    \end{adjustbox}
    \end{center}
\end{table}
Notice how recall is non-decreasing as a function of $k$, while precision can fluctuate up and down.
By varying $k$ between 1 and the total number of detections, we can plot a precision vs. recall curve (see Figure~\ref{fig:precision-recall}).
The precision-recall curve (orange) is made non-increasing by taking the \textit{interpolated} precision (green), defined as $p_{interp}(r)=\max_{r'\geq r} p(r)$.\footnote{Interpolation is a natural thing to do because it means that for a minimum recall requirement, there exists a confidence-threshold which results in a detector with better precision if we allow the recall to be higher than that threshold, which is never a detrimental.} The average precision is defined as the area under that curve. The exact way the area is computed depends on the specific benchmark.

For PASCAL VOC's 2007 test set~\citep{pascal}, which is used for FSOD evaluation, the main IoU-threshold used is 0.5 and the area under the curve is approximated by sampling and averaging the interpolated precision at 11 points $\lbrace 0, 0.1, 0.2, \dots, 1\rbrace$. IoU-thresholds of 0.75 and COCO-style mAP are also sometimes reported~\citep{tfa}.
For MS COCO, the area under the interpolated precision curve is computed exactly by adding up the area of each rectangle (see Figure). COCO-style mAP is defined by averaging the mAP at different thresholds from 0.5 to 0.95 in increments of 0.05. For COCO, it is common to report mAP scores for objects of different sizes: small, medium and large~\citep{tfa,fsdetview}. Additionally, mAP50 and mAP75 are often provided~\citep{tfa,attentionfsod,metadetr}, which are computed by computing the exact area under the curve for IoU-thresholds of 0.5 and 0.75.
Generally in FSOD, it is also common to report mAP separately for base and novel classes, denoted as bAP and nAP~\citep{tfa}. Most works put greater emphasis on the nAP~\citep{metayolo,metarcnn}, though some claim that maximizing bAP is also important to avoid catastrophic forgetting~\citep{retentivercnn,tfa}.
These metrics and datasets give us a comprehensive overview of how different models perform for few-shot object detection.
Note that for LVIS v0.5, the metrics are the same as for MS COCO, and reported separately for frequent, common and rare objects. However, we do not recommend following the TFA splits for LVIS due to the issues outlined in Section~\ref{sec:fsod-dataset-discussion}.

\subsection{Few-Shot Object Detection Methods}

We review several FSOD methods from the literature. In our description, we assume that backbones have already been pretrained on ImageNet. We summarize most of these  methods in Table~\ref{tab:fsod}.\footnote{We only included methods evaluated on at least one of the dominant FSOD benchmarks/splits.}

\newcommand{\ft}{\colorbox{Emerald!30!white}{finetuning}}
\newcommand{\Ft}{\colorbox{Emerald!30!white}{Finetuning-only baselines}}
\newcommand{\proto}{\colorbox{RedOrange!30!white}{prototype}}
\newcommand{\Proto}{\colorbox{RedOrange!30!white}{Prototype-based methods}}
\newcommand{\modulation}{\colorbox{Fuchsia!30!white}{modulation}}
\newcommand{\Modulation}{\colorbox{Fuchsia!30!white}{Modulation-based methods}}
\newcommand{\addon}{\colorbox{gray!30!white}{add-on}}
\newcommand{\Addon}{\colorbox{gray!30!white}{Add-on methods}}

\begin{table*}[!t]
\centering
\caption{\textbf{Few-Shot Object Detection methods with results on PASCAL VOC and MS COCO.} Methods are categorized as \ft-only, \proto-based, and \modulation-based. TIP is a general add-on strategy for two-stage detectors. Faster RCNN+FT numbers are from TFA~\citep{tfa}. RepMet and Attention-FSOD numbers are from Meta-DETR~\citep{metadetr}. Methods are sorted by MS COCO 30-shot nAP. \\Find the most up-to-date table at~\url{https://github.com/gabrielhuang/awesome-few-shot-object-detection}}
\label{tab:fsod}
\begin{adjustbox}{width=0.8\textwidth}
\begin{tabular}{llrrrrrrrr}
\toprule
  &   & \multicolumn{3}{c}{VOC TFA-split (nAP50)} & \multicolumn{3}{c}{VOC Kang-split (nAP50)} & \multicolumn{2}{c}{MS COCO (nAP)} \\
  \cmidrule(r){3-5} \cmidrule(lr){6-8} \cmidrule(l){9-10}
Name & Type & \thead{1-shot} & \thead{3-shot} & \thead{10-shot } & \thead{1-shot } & \thead{3-shot } & \thead{10-shot } & \thead{10-shot} & \thead{30-shot} \\
\midrule
LSTD~\citep{lstd} & \ft & - & - & - & 8.2 & 12.4 & 38.5 & - & - \\
RepMet~\citep{repmet} & \proto & - & - & - & 26.1 & 34.4 & 41.3 & - & - \\
Meta-YOLO~\citep{metayolo} & \modulation & 14.2 & 29.8 & - & 14.8 & 26.7 & 47.2 & 5.6 & 9.1 \\
MetaDet~\citep{metadet} & \modulation & - & - & - & 18.9 & 30.2 & 49.6 & 7.1 & 11.3 \\
Meta-RCNN~\citep{metarcnn} & \modulation & - & - & - & 19.9 & 35.0 & 51.5 & 8.7 & 12.4 \\
Faster RCNN+FT~\citep{tfa} & \ft & 9.9 & 21.6 & 35.6 & 15.2 & 29.0 & 45.5 & 9.2 & 12.5 \\
ACM-MetaRCNN~\citep{acmmetarcnn} & \modulation & - & - & - & 31.9 & 35.9 & 53.1 & 9.4 & 12.8 \\
TFA w/fc~\citep{tfa} & \ft & 22.9 & 40.4 & 52.0 & 36.8 & 43.6 & 57.0 & 10.0 & 13.4 \\
TFA w/cos~\citep{tfa} & \ft & 25.3 & 42.1 & 52.8 & 39.8 & 44.7 & 56.0 & 10.0 & 13.7 \\
Retentive RCNN~\citep{retentivercnn} & \ft & - & - & - & 42.0 & 46.0 & 56.0 & 10.5 & 13.8 \\
MPSR~\citep{mpsr} & \ft & - & - & - & 41.7 & 51.4 & 61.8 & 9.8 & 14.1 \\
Attention-FSOD~\citep{attentionfsod} & \modulation & - & - & - & - & - & - & 12.0 & - \\
FsDetView~\citep{fsdetview} & \modulation & 24.2 & 42.2 & 57.4 & - & - & - & 12.5 & 14.7 \\
CME~\citep{cme} & \ft & - & - & - & 41.5 & 50.4 & 60.9 & 15.1 & 16.9 \\
TIP~\citep{tip} & \addon & 27.7 & 43.3 & 59.6 & - & - & - & 16.3 & 18.3 \\
DAnA~\citep{dana} & \modulation & - & - & - & - & - & - & 18.6 & 21.6 \\
DeFRCN~\citep{defrcn} & \proto & - & - & - & 53.6 & 61.5 & 60.8 & 18.5 & 22.6 \\
Meta-DETR~\citep{metadetr} & \modulation & \mr{35.1} & \mr{53.2} & \mr{62.0} & - & - & - & \mr{19.0} & \mr{22.2} \\
DETReg~\citep{detreg} & \ft & - & - & - & - & - & - & \mr{25.0} & 30.0 \\
\bottomrule
\end{tabular}
\end{adjustbox}

\end{table*}

\subsubsection[title]{\Ft~\footnote{Use colors for quick reference to Table~\ref{tab:fsod}.}}

Finetuning-only methods generally start from a traditional object detector such as Faster-RCNN~\citep{fasterrcnn} with only minor architecture modifications. They do base training on many base class examples, then do few-shot finetuning on a support set containing both base and novel classes. The rationale behind this two-step process is to deal with the extreme imbalance between the base and novel classes, and to avoid overfitting on the novel classes.

LSTD~\citep{lstd} proposed the first finetuning strategy for FSOD. A hybrid SSD/Faster-RCNN ``source'' network is trained on the source classes, then at finetuning time, its weights are copied into a ``target'' network, except for the classification layer, which is randomly initialized, and finetuned on the target classes.\footnote{Slightly different from the FSOD framework presented in Section~\ref{sec:fsodframework}, the authors exclusively consider a cross-dataset scenario; therefore, target classes may or may not include sources classes.} Additionally, the authors propose to regularize the finetuning stage by penalizing the activation of background features with $L_2$ loss (Background Depression), and another ``Transfer-Knowledge'' loss which pulls target network predictions closer to the source network predictions.
Subsequently, TFA or Frustratingly Simple Few-Shot Object Detection~\cite{tfa} showed that even a simple finetuning approach with minimal modifications to Faster R-CNN can actually yield competitive performance for FSOD. TFA replaces the fully-connected classification heads of Faster R-CNN with \textit{cosine} similarities; the authors argue that such feature normalization leads to reduced intra-class variance, and less accuracy decrease on the base classes. First, a Faster R-CNN with cosine classification heads is trained on the base classes using the usual loss. During few-shot finetuning, new classification weights are randomly initialized for novel classes, appended to the base weights, and the last layers of the model are finetuned on the base+novel classes, while keeping the backbone and RPN frozen. 
MPSR~\citep{mpsr} is also a finetuning approach. They propose an even more scale-aware Faster RCNN by combining the Feature Pyramid Network with traditional object pyramids~\citep{adelson1984pyramid}. After training the model on the base classes, the classification layer is simply discarded, a new classification layer is initialized, and the model does few-shot finetuning on the base+novel classes without freezing any layers.
RetentiveRCNN~\citep{retentivercnn} extend TFA to generalized FSOD, where the goal is to perform well on the novel classes without losing performance on the base classes. They observe a ``massive variation of norms between base classes and unseen novel classes'', which could explain why using cosine classification layers is better than fully connected ones. After training a Faster RCNN on the base classes, they freeze the base RPN and detection branches, and in parallel they introduce new finetuned RPN and detection branches to detect both base and novel classes. They also use a consistency loss, similar in spirit to LSTD's transfer-knowledge loss, to make predictions on base objects more similar in the base/base-and-novel branches.
DETReg~\citep{detreg} use a finetuning approach on the Deformable DETR~\citep{deformabledetr} architecture, and achieve state-of-the-art results on few-shot COCO object detection after proposing a self-supervised strategy for pretraining the detection heads, which is discussed in more depth in Section~\ref{sec:predictiveapproaches}.

\subsubsection{Conditioning-based methods}

For clarity, we will refer to the image to process (to detect objects from) as the \textbf{query image}.
In addition to the query image, conditioning-based methods are also fed with annotated \textbf{support images}, which are reference examples of each class to detect. Each support image generally has a single bounding-box around the object to detect.
In the context of the FSOD framework presented in Section~\ref{sec:fsodframework}, support images are randomly sampled from all base classes during training (step 3), while a predefined (few-shot) set of base and novel images are used during finetuning and evaluation (steps 4 and 5). In this section, we review two types of conditioning-based methods: prototype-based, and modulation-based.

\paragraph{\Proto}

RepMet~\citep{repmet}, which stands for representative-based metric learning, is based on Faster-RCNN with Deformable Feature Pyramid Network (FPN), and learns representatives/prototypes for each object category. At base-training time, RepMet samples several supporting examples for each class, computes their ROI-pooled representations (representatives), and classifies object proposals according to their distance to the representatives. The gradients are propagated through both the proposals and the prototypes.
At few-shot finetuning and evaluation time, representatives for the novel classes are computed, and objects proposals are classified using those new representatives. Optionally, the authors propose to finetune the novel prototypes by maximizing the detection loss on novel objects, which they find beneficial (denoted as ``episode fine-tuning'' in Table 3 of ~\citet{repmet}). 
ACM-MetaRCNN~\citep{acmmetarcnn} have also proposed a baseline which combines Faster R-CNN with prototypical networks by replacing the classification layer with the non-parametric prototypical network equivalent. They investigate this baseline with and without finetuning, and find that it is always beneficial to finetune.

\paragraph{\Modulation\label{sec:modulationbased}}

Modulation-based methods generally compute \textbf{support weights} (also known as \textit{class embeddings}, \textit{weights}, \textit{prototypes}, or \textit{attentive-vectors}) from the support features using a separate \textbf{conditioning branch} (sometimes called \textit{reweighting} module~\citep{metayolo},  \textit{guidance} module~\citep{tip}, or \textit{remodeling} network~\citep{metarcnn}). Each class has its own support weights, which are usually computed by applying global average-pooling to the support features and have a shape $1\times 1\times C$ shape, where $C$ is the number of channels. The support weights then are multiplied channel-wise with the query features to obtain class-specific query features, in a process known as \textbf{modulation} or \textbf{aggregation}. Finally, binary detections are predicted (\textit{object} vs. \textit{background}) on each set of class-specific features, and the results are merged to get multiclass detections. For faster inference, support weights can be precomputed and stored during the finetuning/conditioning step.


\shave{
\begin{figure*}
\centering
\includegraphics[width=0.8\linewidth]{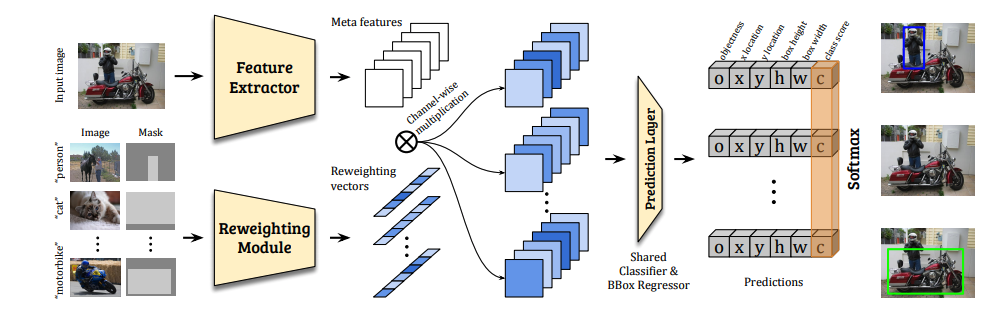}
\caption{\textbf{Meta-YOLO}, a modulation-based FSOD method.}
\label{fig:conditioning-based}
\end{figure*}
}


To the best of our knowledge, one of the first modulation-based method was Meta-YOLO~\cite{metayolo}, which is also the work that introduces the standardized FSOD splits\shave{ (see \textbf{Figure~\ref{fig:conditioning-based}})}. Meta-YOLO uses both conditioning and finetuning. The model uses a feature extractor module to obtain the image features of the query image, and a reweighting module that extracts features from the bounding boxes of the support images to obtain reweighting vectors. The object mask binary matrices are added as an extra layer to the RGB images to form 4-channel images which are fed to the reweighting module. These vectors are used to condition the image features which allows the prediction layer to output the bounding boxes corresponding to the classes represented by those vectors. During base training, the two modules and the prediction layer are trained on the base classes. During few-shot finetuning, the model is finetuned using the $K$ support examples per class, from the base and novel classes.\footnote{Since only $K$ labeled bounding boxes are available for the novel classes, to balance between samples from the base and novel classes, only $K$ boxes are included for each base class} At few-shot evaluation time, reweighting vectors for each class are precomputed by averaging the corresponding vectors, and used to modulate the main features.

Since then, several improved conditioning-based methods have been introduced.
Due to using different architectures and design choices, these works employ a diverse range of modulation strategies, which we discuss below.
In \textbf{Meta-YOLO}~\citep{metayolo}, which is based on the single-stage detector YOLO, the features $f_{qry}$ output by the backbone are directly multiplied channel-wise by the class embeddings $f_{cls}$, resulting in the modulated features $[f_{qry} \bigotimes f_{cls}]$. 
In \textbf{Meta-RCNN}~\citep{metarcnn} which is based on the two-stage detector Faster RCNN, the features are only multiplied after they have been pooled with RoIAlign; the consequence is that the Region Proposal Network (RPN) is agnostic to the object category. 
In \textbf{ACM-MetaRCNN}~\citep{acmmetarcnn}, also based on Faster RCNN, the feature maps are multiplied twice by the class prototypes: before running the RPN, and after pooling the features of each box, which means the RPN may produce different region proposals for each class.
In \textbf{Attention-FSOD}~\citep{attentionfsod}, based on Faster RCNN, the query features are convolved channel-wise with support features (used as kernels) before feeding them to the RPN. In practice the authors find $1\times 1$ support features to be optimal, and the convolution reduces to a channel-wise multiplication. The proposals are fed to a relational detection head, which classifies objects using matching scores by  comparing query features with support features. Additionally, the authors explore three ways to model support-to-query relations in the detection head: globally, locally, and patch-wise. They find it beneficial to use all three types of relation heads.
In \textbf{Fully Guided Network (FGN)}~\citep{fgn}, based on Faster RCNN, the support features are global-average-pooled and averaged for each class to obtain the class-attentive vectors. Then, the query features are multiplied channel-wise with the class-attentive vectors and fed to the RPN to obtain class-specific proposals, which are aggregated between the different classes. Finally, in the relational detection head, the aligned query features and $N$ support class averages are concatenated altogether and fed into a MLP to obtain box and multiclass predictions. 
Relational detection heads~\citep{relationnet} have the ability to jointly predict boxes for all classes, which differs from other conditioning-based approaches which predict boxes independently for each class by relying on class-specific modulated features.
In \textbf{FsDetView}~\citep{fsdetview}, the query features are modulated by the support weights after ROI pooling. Instead of simply multiplying the features together, the authors propose to also concatenate and subtract them, resulting in the modulated features $[f_{qry} \bigotimes f_{cls}, f_{qry} - f_{cls}, f_{qry}]$.
\textbf{Meta-DETR}~\citep{metadetr} adapts FsDetView's modulation strategy to the DETR architecture~\citep{detr}. At the output of the backbone, both support and query features are fed to the transformer encoder, then the query features are multiplied with global-average-pooled support vectors, and fed to the transformer decoder to make binary predictions.

\paragraph{Necessity of finetuning}

Only a few FSOD methods such as \citet{li2020one} or 
Attention-FSOD~\citep{attentionfsod} present themselves as methods that do not require finetuning.
However, we should note that most if not all of the conditioning-based methods presented in the previous section could technically be used without finetuning on base+novel classes, by directly conditioning on the support examples at few-shot evaluation time.
In practice, most works find it beneficial to finetune, and in fact many of the conditioning-based methods reviewed above do not even report numbers without finetuning.
For instance, ACM-MetaRCNN, a conditioning-based model, finetune their model, except for 1-shot and 2-shot on PASCAL VOC, where they do not finetune ``to avoid overfitting''. 
Even Attention-FSOD~\citep{attentionfsod}, which claims to be usable without finetuning, achieves its best performance after finetuning (see for instance Table~\ref{tab:fsod}).

\subsubsection{\Addon}

Some FSOD methods do not propose a specific architecture, but instead propose add-on tricks that can be combined with many of the existing FSOD methods to boost performance. 
For instance, Transformation Invariant Principle (TIP)~\citep{tip} is a regularization strategy based on data augmentation transformations, which can be applied to any two-stage FSOD method. Specifically, TIP proposes to minimize several consistency losses: the guidance vectors (class weights) for a support object (first view) and its transformed version (second view) should be close in feature space (the authors find the L2 distance to give best results). Additionally, TIP pools features from one view using proposals generated from another view; the resulting detections are used to compute another detection loss (regression and classification). 
%

\section{Self-Supervised pretraining \label{sec:ss}}

\colorlet{predictive}{YellowOrange!30!white}
\colorlet{contrastive}{Aquamarine!30!white}
\colorlet{byol}{Thistle!30!white}
\colorlet{clustering}{Tan!30!white}

%

Until recently, the standard approach in deep object detection was to pretrain the backbone on supervised ImageNet~\cite{deng2009imagenet} classification. This still holds for modern iterations of two-stage detectors such as Faster R-CNN~\cite{fasterrcnn} -- as per its \texttt{detectron2} implementation -- as well as one-stage detectors such as YOLOv3~\cite{yolov3}, SSD~\cite{ssd}, RetinaNet~\cite{retinanet} and recent transformer-based detectors like DETR~\cite{detr} and Deformable-DETR~\cite{deformabledetr}. 

Self-supervised pretraining has emerged as an effective alternative to supervised pretraining where the supervision comes from the data itself. The key idea is to automatically generate labels from unlabeled data, and learning to predict those labels back. This process is known as solving a pretext task. For instance, a common pretext task is to predict the relative position of two random crops from the same image~\citep{pathak2016context}. 
This broad definition could potentially include many unsupervised methods such as VAEs~\citep{kingma2013auto} and GANs~\citep{goodfellow2020generative,huang2017parametric} but in practice the self-supervised term is used for methods for which the pretext task differs from the downstream task~\citep{moco,simclr,byol}. Some language models such as word2vec~\citep{mikolov2013efficient} are also considered to be self-supervised.


Starting with SimCLR~\cite{simclr} and MoCo~\cite{moco}, people have experimented initializing object detection backbones with unsupervised representations learned on ImageNet (or COCO) instead of supervised ones.
Since the pretext tasks are fairly general, there is the hope that unsupervised representations might generalize better to downstream tasks than classification-based ones. Recent works~\citep{soco,detreg,insloc,densecl} which we will refer to as \textit{self-supervised object detection} methods go beyond backbone-pretraining by also pretraining the detection heads specifically for object detection.

In Section~\ref{sec:ssc} we review self-supervised \textit{classification} methods; then in Section~\ref{sec:issues-ssc} we discuss their limitations for initializing object detection backbones; finally in Section~\ref{sec:ssod} we review self-supervised \textit{object detection} approaches, which unlike the previous methods, are specifically tailored to object detection.

\subsection{Image-level Backbone Pretraining \label{sec:ssc} \label{sec:pretrainingbackbone}}
In the image classification domain, self-supervised learning has emerged as a strong alternative to supervised pretraining, especially in domains where images are abundant but annotations are scarce~\cite{manas2021seasonal}. 
Self-supervised classification methods are not limited to classification, as the learned feature extractor can used to initialize the backbone of common object detection architectures.
We categorize backbone pretraining strategies into constrastive, clustering-based and self-distillative.
We will omit reconstruction~\citep{vincent2008extracting,pathak2016context,zhang2016colorful,zhang2017split} methods and visual common sense~\citep{doersch2015unsupervised,noroozi2016unsupervised,gidaris2018unsupervised} based tasks, as to our knowledge, they have not been used for object detection.

\subsubsection{\colorbox{contrastive}{\mr{Global} Contrastive Learning \xmark} \protect\footnote{Use colors for quick reference to Table~\ref{tab:ssod}. The \xmark~means backbone-only pretraining.} \label{sec:global-contrastive-learning}} These approaches leverage the InfoNCE~\citep{oord2018representation} loss to compare pairs of samples which can be positive pairs or negative pairs. Usually, positive pairs are generated from different \textit{views} (data augmentations) of the same image, while negative pairs are generated from different images.
Contrastive Predictive Coding (CPC) is one of the first approaches to be competitive with respect to supervised classification~\citep{oord2018representation,henaff2020data}. 
In CPC, the goal is to predict a future part of a sequential signal $p(x|c)$ given some previous context of that signal ($c$). Since reconstructing $x$ from $c$ is difficult in high-dimensional spaces, they propose a contrastive objective instead. Given a set of random samples, containing one \textit{positive} sample $x^{+}\sim p(x|c)$, and $N-1$ \textit{negative} samples $x_{1}^{-},\dots,x_{N}^{-}\sim p(x)$ from the ``proposal'' distribution, they propose to learn a function $f_\theta(x,c)$ which minimizes the InfoNCE loss:
%
%
\begin{align}
\mathcal{L_{\text{InfoNCE}}(\theta)} &= - \mathbb{E}_X\left[\log \frac{f_\theta(x^{+}, c)}{\sum_{i} f_\theta(x_i^{-}, c)}\right] \label{loss1}.
\end{align}
The density ratio $f_\theta(x, c)$ can be interpreted as an affinity score, which should be high for the real sample and low for negative samples.

More recent approaches such as momentum contrast (MoCo)~\citep{moco,mocov2,mocov3}, or SimCLR~\cite{simclr,chen2020big} reinterpret the InfoNCE loss in a different context. Given a reference image $x_0$, positive samples $x^{+}\sim p(x|x_0)$ are generated using data augmentation on $x_0$, and negative samples $x_{1}^{-},\dots,x_{N}^{-}\sim p(x)$ are other images sampled from the dataset. The InfoNCE loss becomes:
\begin{align}
\mathcal{L_{\text{InfoNCE}}(\theta)} &= - \mathbb{E}_X\left[\log \frac{f_\theta(x^{+}, x_0)}{\sum_{i} f_\theta(x_i^{-}, x_0)}\right] \label{loss2}.
\end{align}
The goal is to learn representations which maximize the affinity $f_\theta(x, x_0)$ between different data-augmentations (\textbf{views}) of the same image, and minimize the affinity between different images.

While both MoCo and SimCLR find crucial to use a large set of negative examples, they differ in their approach to obtain them. SimCLR uses a large batch size, while MoCo uses smaller batches but stores the embeddings in a queue, which is used to retrieve negative examples. To prevent drift between queued embeddings and positive examples, MoCo encodes negative examples with an exponential moving average of the weights.

Most self-supervised pre-training methods in the literature aim to learn a global image representation to transfer to a given downstream task~\citep{moco,mocov2,mocov3,simclr,swav,dino}. However, global image representations  might not be optimal for dense prediction tasks such as detection and segmentation. In order to bridge the gap between self-supervised pre-training and dense prediction, \citet{pinheiro2020unsupervised} proposed VADeR, a pixel-level contrastive learning task for dense visual representation. Different from global contrastive learning approaches such as SimCLR, VADeR uses an encoder-decoder architecture. Then, given the output feature maps for a positive sample, an augmented sample, and a negative sample, the InfoNCE loss is applied between the decoder's pixel features rather than being applied to the average of the decoder's output. As a result, VADeR achieves encouraging results when compared to strong baselines in many structured prediction tasks, ranging from recognition to geometry. 

\subsubsection[title]{\colorbox{clustering}{Clustering-Based \xmark}~\footnote{Use colors for quick reference to Table~\ref{tab:ssod}. Checkmarks \cmark~and \xmark~indicate whether the detection heads are pretrained.}}
Clustering-based methods rely on unsupervised clustering algorithms to generate pseudo-labels for training deep learning models~\citep{xie2015unsupervised,yang2016joint}. A key idea is to alternate between clustering learned representations, and using the predicted cluster assignments to improve representations in return.
\citet{caron2018deep,caron2019unsupervised} show that k-means cluster assignments are an effective supervisory signal for learning visual representations. \citet{asano2019self} show that cluster assignments can be solved as an optimal transport problem. Based on previous approaches, Swapping Assignments between multiple Views of the same image (SwAV)~\citep{swav} was proposed. SwAV attempts to predict the cluster assignments for one view from another view (data augmentation) of the same image. It uses the Sinkhorn-Knopp algorithm for clustering~\citep{cuturi2013sinkhorn}, which has previously been explored for recovering labels in few-shot classification~\citep{huang2019few}, and has good properties such as quick convergence and differentiability~\citep{cuturi2013sinkhorn}. SwAV avoids trivial solutions where all features collapse to the same representation by using the appropriate amount of entropy regularization~\citep{swav}.
 
\subsubsection{\colorbox{byol}{Knowledge Self-distillation (BYOL) \xmark}} 
Self-distillative approaches such as Bring Your Own Latent (BYOL)~\citep{byol} and mean teacher~\citep{tarvainen2017mean} move away from contrastive learning by maximizing the similarity between the predictions of a teacher and a student model. 
The student model is optimized using \mr{SGD}, while the teacher model is instantiated as an exponential moving average of the student weights~\citep{byol}. In order to prevent them from collapsing to the same representation, \citet{byol} found it helpful to use other tricks such as softmax sharpening and recentering.
Subsequent approaches such as DINO~\citep{dino} and EsViT~\citep{esvit} leverage vision transformers (ViT)~\citep{dosovitskiy2020image} instead of residual networks, following the trend of using self-supervision in natural language processing~\citep{vaswani2017attention,radford2018improving,brown2020language}. 
They divide the input image into a grid of small patches ($8\times 8$ pixels for DINO) and feed them to a ViT. The last feature map, which could be used as a dense representation, is average-pooled into a single vector and compared between teacher and student models using cross-entropy.
The main difference between the two is that EsViT uses a two-stage architecture and a part-based loss. The first improvement reduces the amount of image patches in the second stage, which makes the model more efficient. The second improvement introduces an additional loss besides DINO's teacher-student loss that encourages matching regions across multiple views to match their respective student and teacher representations. 

\begin{figure}[t]
\centering
\includegraphics[width=\linewidth]{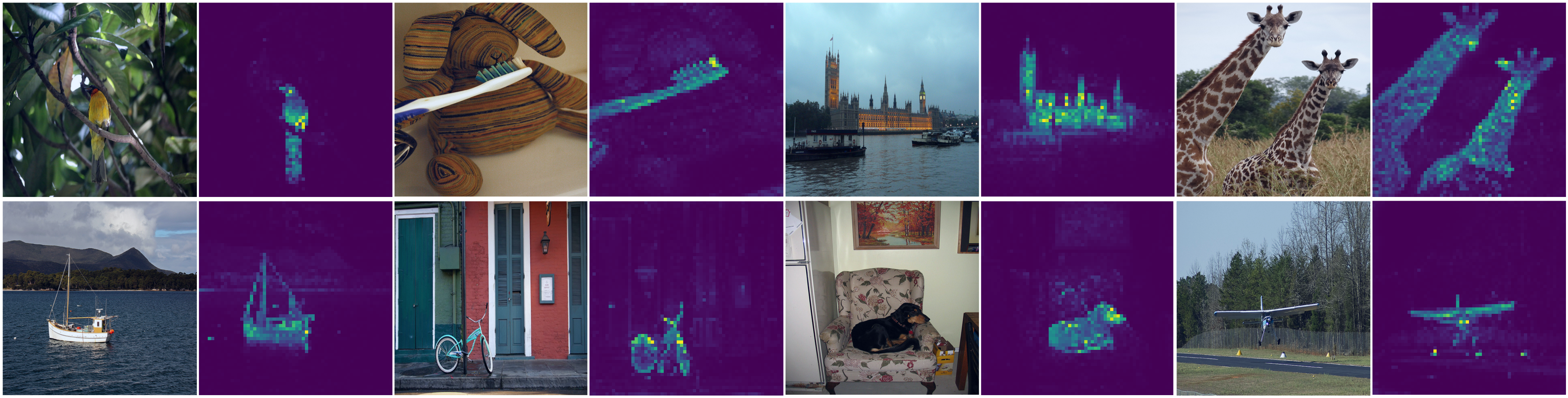}
\caption{\textbf{DINO's attention maps.} Since DINO is based on a visual transformer, the attention maps corresponding to the \texttt{[CLS]} token can be plotted. Despite being trained with no supervision, different attention heads are found to segment different objects. Source:~\citet{caron2019unsupervised}. \label{fig:dino}}
\end{figure}

Interestingly, the authors of DINO show that semantic segmentation masks naturally emerge from the attention masks of the visual transformer (see Figure~\ref{fig:dino}). This is very surprising giving that the model was trained with no supervision whatsoever. This suggests that combining dense pretext tasks such as VADeR~\citep{pinheiro2020unsupervised} and DINO~\citep{dino} with attention-based models such as transformers could improve the transferability of self-supervised learning methods to dense downstream tasks.

\subsection{Issues of Combining Self-supervised Classification with Detection\label{sec:issues-ssc}}

Conceptually, there are some issues with transferring classification-based representations to object detectors, whether the representations were supervised or self-supervised.

\subsubsection{Untrained Detection Heads} The first issue is with untrained detection heads, due to the architecture mismatch. For instance, it is a common practice to combine Resnet-50 and Resnet-101 backbones with a Feature Pyramid Network (FPN). However, the FPN is initialized from scratch and does not generally benefit from pretraining (\mr{though \citet{pinheiro2020unsupervised} propose to pretrain the FPN)}. The same goes for the Region Proposal Network (RPN) and the detection regression and classification heads, in the case of Faster RCNN; and for the encoder and decoder, in the case of DETR-style architectures.

\subsubsection{Task Mismatch} Secondly, ImageNet top-1 classification accuracy does not necessarily correlate with object detection performance. While certain properties such as translation and scale invariance are desirable for classification, they might actually hinder object localization~\citep{newell2020useful,xiao2020should}. 
Classification-based representations such as MoCo~\cite{moco} or SimCLR~\cite{simclr} might discard spatial information that is useful for localization, because they are irrelevant for solving the pretraining tasks.
Moreover, data augmentation strategies such as random cropping and jittering may introduce undesirable invariances into the network.
In fact, \citet{insloc} show that MoCo can perform better on object detection than BYOL and SwAV, despite having worse classification performance. 
\mr{Identifying whether the performance bottleneck comes from localization or classification errors is as hard of a problem as evaluating object detection itself.
Metrics like mAP cannot disentangle localization and classification errors—AP is computed per-category, so if class predictions are wrong they will impact the localization error. mAP with higher IoU threshold put more emphasis to precise localization, but they are still contingent on having correct class assignments.}

\subsection{\mr{Object Detection Pretraining} \label{sec:ssod}\label{sec:pretrainingdetection}}

Self-supervised object detection approaches attempt to remedy those issues by pretraining the object detection pipeline on a variety of unsupervised pretext tasks.

\begin{table*}
    \centering
    \caption{Comparison of Self-Supervised Object Detection Methods pretrained on unlabeled ImageNet. The check marks \cmark~and \xmark~refer to whether the methods pretrain both the backbone and detection heads vs. only the backbone.
    }
    \label{tab:ssod}

    \begin{adjustbox}{width=0.8\textwidth}

\begin{tabular}{lccccrrll}
\toprule
Name & \thead{Pretrains\\Detector} & Loss & \thead{View\\Matching} & \thead{Region\\Proposal\\Mechanism} & \thead{Pascal\\AP50} & \thead{COCO\\AP} & \thead{Best\\Backbone} & \thead{Object\\Detector} \\
\midrule
DETReg~\citep{detreg} & \cmark & \colorbox{predictive}{predictive \cmark} & - & selective search & 83.3 & 45.5 & R50 & Def.DETR \\
SoCo~\citep{soco} & \cmark & \colorbox{byol}{BYOL \cmark} & crop & selective search & 83.8 & 44.3 & R50-FPN & F-RCNN \\
InsLoc~\citep{insloc} & \cmark & \colorbox{contrastive}{contrastive \cmark} & crop & random crop & 83.0 & 43.3 & R50-C4 & F-RCNN \\
UP-DETR~\citep{updetr} & \cmark & \colorbox{predictive}{predictive \cmark} & - & random crop & 80.1 & 42.8 & R50-C4 & DETR \\
ReSim~\citep{resim} & \cmark & \colorbox{contrastive}{contrastive \cmark} & sliding window & random crop & 83.1 & 41.4 & R50-FPN/C4 & F-RCNN \\
DenseCL~\citep{densecl} & \cmark & \colorbox{contrastive}{contrastive \cmark} & feature & random crop & 82.8 & 41.2 & R50-FPN/C4 & F-RCNN \\
\midrule
VaDer~\citep{pinheiro2020unsupervised} & FPN-only & \colorbox{contrastive}{contrastive \xmark} & feature & - & - & 39.2 & R50-FPN & F-RCNN \\
\midrule
EsViT~\citep{esvit} & \xmark & \colorbox{byol}{BYOL \xmark} & image & - & - & 46.2 & Swin-ViT & F-RCNN \\
DETcon~\citep{detcon} & \xmark & \colorbox{contrastive}{contrastive \xmark} & mask & grid/FH/MCG & 82.8 & 43.4 & R50-FPN & F-RCNN \\
BYOL~\citep{byol} & \xmark & \colorbox{byol}{BYOL \xmark} & image & & 81.0 & 42.3 & R50-FPN/C4 & F-RCNN \\
DI\xmark~\citep{dino} & \xmark & \colorbox{byol}{BYOL \xmark} & image & - & - & - & ViT/R-50 & - \\
SwAV~\citep{swav} & \xmark & \colorbox{clustering}{clustering \xmark} & image & - & 77.4 & 42.3 & ResNet & F-RCNN \\
MoCo~\citep{mocov2} & \xmark & \colorbox{contrastive}{contrastive \xmark} & image & - & 82.5 & 41.7 & R50-FPN/C4 & F-RCNN \\
SimCLR~\citep{simclr} & \xmark & \colorbox{contrastive}{contrastive \xmark} & image & - & 81.9 & 39.6 & ResNet & F-RCNN \\
\bottomrule
\end{tabular}
\end{adjustbox}
\end{table*}

\subsubsection{\colorbox{predictive}{Predictive approaches \cmark \label{sec:predictiveapproaches}}}

Predictive approaches such as UP-DETR~\cite{updetr} and DETReg~\cite{detreg} pretrain the detection heads of DETR by making them re-predict the position of automatically generated ``ground-truth'' crops. These crops are generated either randomly for UP-DETR or using Selective Search~\citep{uijlings2013selective} for DETReg, a training-free heuristic based on iteratively merging regions with similar colors, textures, and other local characteristics. \footnote{Note that Selective Search used to be a popular training-free heuristic for generating high-recall low-precision region proposals, and was used in RCNN~\cite{rcnn} and Fast RCNN~\cite{fastrcnn} before it was replaced by a trained Region Proposal Network (RPN). }

To pretrain DETR on unsupervised images, the multi-class classification heads (see Figure~\ref{fig:detr}) which would normally predict the object category or \textit{background} are replaced with a \textit{binary} classification head.
In DETReg, the goal is to detect the top proposals generated by Selective Search, as if they were ground-truth foreground objects. The usual DETR loss is used, except that the matching cost used to compute correspondences between ground-truth boxes and detections is a function of the predicted binary labels and locations -- instead of ground truth and predicted multiclass labels.
In UP-DETR, the goal is to predict back the positions of the random crops. Specifically, the transformer decoder is conditioned on the random crops by adding their corresponding features to the decoder input (see \textit{object queries} in Figure~\ref{fig:detr}). This is done by partitioning the object queries into $K$ groups,\footnote{The way UP-DETR matches predictions and ground-truth from different groups instead of matching only within the same groups might not necessarily be an intended feature, but rather a consequence of building on top of existing DETR code. In practice, this does not make much difference as the groups are matched correctly (personal communication with the authors).} and adding a different random crop to each group. The loss is computed by finding the optimal matching between the predicted boxes and the ``ground-truth'' random crops using the Hungarian algorithm~\citep{munkres1957algorithms}, where the cost of matching two boxes is a function of their location and predicted binary label.

On top of the DETR loss, an additional reconstruction loss is used to force the decoder transformer to reconstruct its input. DETReg uses a simple L1 loss $\L_{rec}(z_i,z_j)=||z_i-z_j||_{1}$, while UP-DETR uses cosine similarity $\L_{rec}(z_i,z_j)=\bigl\lVert z_i/\lVert z_i\rVert-z_j/\lVert z_j\rVert \bigr\rVert_{1}$.
Also, an interesting by-product of UP-DETR is that it learns a conditioning branch, which can be reused directly for one-shot object detection by replacing the random crops with support images. The paper provides some results on PASCAL VOC~\cite{updetr}.\footnote{Not reported in Table~\ref{tab:ssod} due to using different splits.}

\subsubsection{\colorbox{contrastive}{\mr{Local Contrastive Learning \cmark}}}


\mr{Unlike global contrastive methods (Section~\ref{sec:global-contrastive-learning}) which contrast global representations at the image level, local contrastive methods contrast backbone representations locally, either at the \textit{feature} or \textit{crop} level, with the hope of learning location-aware representations. Some of them, such as InsLoc~\citep{insloc}, also pretrain detection heads.}

In InsLoc~\citep{insloc}, features for each crop are computed using RoIAlign~\citep{fasterrcnn}, then transformed by the RoI heads to a single $1\times 1\times d$ vector. Positive pairs are generated by randomly cropping two views of the same image, while negative pairs are generated by using different images. Specifically in InsLoc, positive pairs are generated by pasting two views of the same object (the foreground) at random locations and scales onto other images of the dataset (the background). The authors also introduce a cut-and-paste scheme in order to force the receptive field of RoIAlign to ignore distractor features outside the bounding box.
 
%
In ReSim~\citep{resim}, two overlapping crops are generated from two different views of the same image. Then, a sliding window is moved across the overlapping area, and the pooled representations are compared at each of the final convolutional layers. Positive pairs consist of aligned sliding windows across two views of the same image, while negative pairs either consist of unaligned sliding windows, or sliding windows from two different images.

In DenseCL~\citep{densecl}, instead of using spatial correspondence, positive pairs are generated by matching each feature from one view to the feature with highest cosine similarity in another view of the same image. 
Negative examples are simply features from different images. Additionally, the authors combine this dense loss with a global MoCo-style loss, which they claim is necessary to bootstrap correct correspondences.

\subsubsection{\colorbox{byol}{Self-distillative approaches (BYOL) \cmark}}

\newcommand{\RoIAlign}{\textrm{RoIAlign}}

Self-distillative (BYOL-based) approaches such as SoCo~\cite{soco} depart significantly from contrastive approaches as there is no need for negative examples. Selective object COnstrastive learning (SoCo) builds on top of BYOL~\cite{byol} and pretrains both the backbone, feature pyramid, and RoI heads of a FPN-based Faster RCNN by training two networks simultaneously. The ``student'' network $f_\theta$ uses \mr{SGD} to copy the feature maps of a ``teacher'' network $f_\xi$, which is an exponential moving average of the student network, and the student network is optimized using \mr{SGD}. 
For a given image, object proposals (denote the bounding boxes $b$) are generated unsupervisedly using Selective Search~\citep{uijlings2013selective}. Then, two views $V_1,V_2$ are generated and respectively fed into student and teacher FPNs to get feature maps $v_1,v_2$. Box features are computed for each bounding box $b$ by pooling $v_1,v_2$ with RoIAlign and passing them to the RoI heads:
$$h_1=f_\theta^H(\RoIAlign(v_1, b), \quad h_2=f_\xi^H(\RoIAlign(v_2, b).$$
The box features $h_1,h_2$ are then projected to obtain latent embeddings $e_1,e_2$, and their cosine similariy is minimized
$$ \mathcal{L}(\theta) = - \frac{ \langle e_1, e_2 \rangle}{ ||e_1||_2 \cdot ||e_2||_2 }.$$
In practice, SoCo introduces several other tricks, such as using more than two views and resizing them at multiple scales, jittering the proposed box coordinates, and filtering proposals by aspect ratio~\cite{soco}.

\subsection{Comparison of Self-Supervised Object Detection Methods}

In Table~\ref{tab:ssod}, we review the self-supervised object detection methods discussed previously , and report their performance on PASCAL VOC and MS COCO object detection (non few-shot). Note that the numbers are not directly comparable in absolute value, due to variations in model architectures, hyperparameters, learning rate schedules, data augmentation schemes, and other implementation details.
Instead, we encourage the reader to dig into the corresponding ablation studies of those works.

\shave{
This table contains methods specifically geared towards object detection as presented in Section~\ref{sec:pretrainingdetection}, which train the backbone (``Pretrains Detector: Yes''), and general purpose representations as presented in Section~\ref{sec:pretrainingbackbone} which only pretrain the backbone on top of which an object detector was fitted a posteriori, often by a subsequent work (``Pretrains Detector: No''). For instance, most of the numbers for DenseCL, BYOL, DETcon, MoCo, SimCLR and SwAV are taken from the SoCo paper~\citep{soco}.
VaDer is in between, as it does not pretrain detection heads but does pretrain a feature pyramid network (FPN) alongside the backbone.
}

The methods can be categorized into four types of losses: self-distillative (\colorbox{byol}{BYOL}), predictive (\colorbox{predictive}{predictive}), constrastive (\colorbox{contrastive}{contrastive}) and clustering-based (\colorbox{clustering}{clustering}). The check marks \cmark and \xmark refer to whether these methods also pretrain the detection heads. 
\textbf{View Matching} refers to the way different views of the same image are matched. From most global to most local: image $>$ cropmask $>$ sliding window $>$ feature.
\textit{Region proposal mechanism} describes how unsupervised regions are generated for the purpose of pretraining (not to be confused with the candidate proposals in two-stage detectors). DETcon generates masks from: ``grid'' a fixed-size grid, ``FH'' the Felzenszwalb-Huttenlocher algorithm~\citep{fh}, or ``MCG'' Multiscale Combinatorial Grouping~\citep{mcg}.
``R50-FPN'' means ResNet-50 with Feature Pyramid Network. ``R50-C4'' means using ResNet-50's C4 layer. ``ViT'' is the visual transformer~\citep{dosovitskiy2020image}. ``Swin'' is a type of hierarchical visual transformer~\citep{swin}. ``F-RCNN'' stands for Faster R-CNN, ``Def. DETR'' for deformable DETR.

Many FSOD methods~\citep{densecl,resim,byol,moco} are found to perform better using multi-scale features with a FPN for MS COCO, but with single-scale C4 features for PASCAL VOC, which may be a consequence of its limited size.

\section{Takeaways \& Trends\label{sec:takeaways}}

We discuss our main takeaways and forecasted trends from this survey.

\subsection{Finetuning is a strong baseline} Almost every few-shot object detection method we have reviewed finetunes on the novel classes. 
This is the case even for conditioning-based methods, which could technically be used without finetuning by conditioning on the support examples, but have been found to benefit from finetuning anyways.
The problem is that finetuning approaches are slower and may require more hyperparameter tuning. This could be a serious obstacle to deploying such methods in the real world.
~In general, we hope to more see competitive finetuning-free methods in the future.

\subsection{Impact of self-supervision for object detection}

It is somewhat surprising that self-supervised object detection pretraining only brings limited improvements for traditional object detection. 
This could be explained by the fact that post-pretraining, the object detector is finetuned on the labeled dataset, which could render self-supervision redundant. It could also be that current experiments are mainly limited to ImageNet and MS COCO pretraining, whilst self-supervision could potentially benefit from larger unlabeled datasets.
However, the impact of self-supervised pretraining seems to be more significant for few-shot and low-data object detection. In fact, the state-of-the-art FSOD results on MS COCO are from DETReg~\citep{detreg}, a self-supervised object detection method.

\subsection{Using heuristics to generate weak labels}

A general trend in self-supervised learning is to use heuristics to generate weak or noisy labels. Data augmentations are now widely used for generating positive pairs in the context of self-supervised classification~\citep{moco,simclr,byol}. Specifically to object detection, DETReg~\citep{detreg} and SoCo~\citep{soco} have adopted Selective Search~\citep{uijlings2013selective}, for generating crops which are more likely to contain objects. On the other hand, DetCon~\citep{detcon} have explored using the Felzenszwalb-Huttenlocher algorithm and Multiscale Combinatorial Grouping to generate better segmentation masks for feature pooling. 
Since these heuristics come from traditional computer vision, we expect practitioners to continue adapting more of them to improve self-supervised training in the future.
An important question is how such heuristics can be integrated in an iterative bootstrapping procedure: as better representations are learned, it might be worthwhile to gradually replace the initial heuristics with learned and improved ones (e.g., replacing selective search with a learned RPN).
One possible direction of research could be to develop differentiable/learnable versions of these traditional heuristics.

\subsection{Rise of transformers}

Visual transformers have gained increasing traction in object detection, both as backbones and as end-to-end detection heads. 
For using them as backbones, works such as DINO~\citep{dino} have shown that fully unsupervised pretraining of visual transformers can lead to the emergence of object segmentation capabilities. Specifically in their case, the multi-head attention modules learn to segment foreground objects as a byproduct of solving the pretext task, as shown in Figure~\ref{fig:dino}.
More generally, there is growing belief from the study of scaling laws for foundational models that visual transformers can generalize better than ResNets to large scale training~\citep{zhai2021scaling}. Some self-supervised methods, such as EsViT~\citep{esvit}, also rely on recent iterations of visual transformers such as Swin~\citep{swin} to obtain state-of-the-art results.
When using them as detection heads, DETR~\citep{detr} has shown that transformer-based detection heads can be trained end-to-end. In particular, they are capable of making joint predictions and dealing with redundant detections, thus removing the need to rely on heuristics such as non-maximum suppression (NMS). More recent work such as Pix2Seq~\citep{chen2021pix2seq} has shown that object detection can be formulated as a language modeling task. The flexibility that comes with language modeling could blur the line between pure vision tasks (such as object detection) and vision-to-language tasks (such as image captioning or visual question answering), lead to simpler architectures, more end-to-end approaches with less heuristics (such as NMS), and pave the way to foundational models for vision and language tasks.

\subsection{Problems with current evaluation procedures\label{sec:takeaways-eval}}

Comparisons such as Table~\ref{tab:fsod} and Table~\ref{tab:ssod} should only be used to get a general idea of the performance of those systems. The numbers themselves often not directly comparable, due to variations in backbone architecture, use of multi-scale features (FPN), varying detection architectures, types of data augmentations used, learning rate scheduling, or even things as trivial as input image size resizing.

\subsubsection{Differences in implementation details} In fact, many of the modulation-based FSOD methods we have reviewed in Section~\ref{sec:modulationbased} have quite a similar structure, differing only in the modulation strategy, backbone architecture and object detector used. 
This raises the question of how much of the performance of state-of-the-art methods is owed to new ideas rather than better hyperparameter tuning or using better architectures. 
One way would be to see how much performance we can get with running older methods in newer frameworks, or building an unifying benchmark.

\subsubsection{Issues with data splits} Specifically to the FSOD use of PASCAL VOC, there has been a shift from using Kang's splits to TFA's splits which were introduced later to alleviate the variance problems with Kang's splits. Despite the fact that the two splits can yield wildly different numbers (see for instance the line on TFA w/cos), several works mistakenly mix them up in the same tables~\citep{fsdetview,tip,metadetr}.
More generally, the fact that virtually every FSOD paper -- regular object detection papers too-- trains on the union of training and validation sets (trainval) and uses the test set for hyperparameter tuning can lead to overfitting and overestimating the actual generalization performance.

\subsubsection{Proposed guidelines}
Therefore, we propose the following guidelines for having more comparable results:
\begin{enumerate}
    \item Define and use proper train/val/test splits. Researchers should agree on newer splits or benchmarks, as no single researcher has any incentive to stop overfitting on the test set.
    \item Do not propose benchmarks which are prone to high variance, such as Kang's splits for Pascal or TFA splits for LVIS. Prior work on few-shot classification has consistently provided confidence intervals by averaging results over multiple episodes, and sampling the few-shot training set instead of fixing the instances~\citep{protonet,chen2018closer,matchingnet}.
    \item Standardize implementation details such as image resizing, whitening, and data augmentations. Define standard backbone and detector architectures to be explored for each benchmark.  Results could be presented in two categories: fixed architecture and best architecture. The introduction of Detectron2 has already led to more standardization and code sharing, providing among other things a standard implementation of Faster R-CNN with FPN.\footnote{Despite the valuable efforts of Detectron2 towards standardization and open-sourcing, we did find the framework overwhelming for some use-cases. This resulted in additional difficulties when working with Detectron2-based projects due to the highly abstract nature of the framework. We hope that future frameworks will be more user-centric. For instance, a micro-framework with independently-usable modules might lead to more readable user code.}
    \item Relate new tasks to existing tasks. For instance, the dominant FSOD and few-shot classification frameworks use different terminologies, training and evaluation procedures, and FSOD could have benefited from FSC best practices. We found it necessary to clarify the differences and subtleties in Section~\ref{sec:fscvsfsod}.
\end{enumerate}

\section{Related Tasks\label{sec:related-approaches}}

We briefly discuss other related tasks, which are out of the scope of this survey. 

\subsection{Weakly-supervised object detection}

Image-level and point-level annotations are cheaper and faster to obtain, and noisy image-level labels could even be generated automatically using image search engines. Weakly supervised object detectors are trained using only image-level annotations without requiring bounding boxes~\citep{bilen2016weakly,jie2017deep,tang2018pcl} and could therefore benefit from a larger pool of labels. Many weakly supervised detection methods fall under
multiple-instance learning (MIL)~\cite{MIL} where each image corresponds to a bag of object proposals. A bag is
given the class label based on whether the label exists in the image. \citet{Li2016WeaklySO}
present a two-step approach that includes selecting good object proposals; then training a Faster
RCNN~\cite{fasterrcnn}. \citet{tang2017multiple} use a refinement learning strategy to select good quality proposals. C-MIL~\cite{Wan2019CMILCM} introduces an optimization method to avoid  selecting poor proposals,  C-WSL~\cite{Gao2018cwsl} uses object count
information to obtain the highest scoring proposals, WISE~\cite{laradji2019masks} that uses class activation maps to score proposals  and LOOC~\cite{laradji2020looc}  can be used to detect objects in crowded scenes. Point-level annotations can also be use for object detection like in~\cite{laradji2019instance,laradji2021weakly,laradji2020proposal}, but they require slightly more human effort.

\subsection{Self-supervision using other modalities}

Instructional videos are a natural source of self-supervision, as they contain both speech and visual information. For instance, \citet{amrani2020self} recently proposed using unlabeled narrated instructional videos to learn an object detector by exploiting the correlations between narration and video. They start by extracting video transcripts with an external method. For a given object, they generate positive frames from the temporal period where the object is mentioned, and negative frames from videos that do not mention the object. They extract bounding boxes using Selective Search~\citep{uijlings2013selective}, compute their features using a pretrained backbone, cluster them in feature space, assign a score to each cluster, and filter out the noisiest examples. Finally, they train a detector on the remaining bounding boxes. \citet{laradji2021weakly2} deal with 3D ct scans using self-supervision and weak supervision to train a model to predict regions in the lungs that are infected with COVID. \citet{liu2019soft} used self-supervision by making sure outputs are consistent between different viewpoints and augmentations which in turn helped improve 2D to 3D reconstruction.


\subsection{Low-data and \mr{semi-supervised} object detection}

Low-data object detection (LSOD) and semi-supervised object detection (SemiOD) are closely related to few-shot object detection (FSOD). Instead of having a distinction between base classes (many examples) and novel classes (few-shot), LSOD and SemiOD both assume that the number of examples for all classes is limited— which is generally simulated by considering a fraction of the labels of traditional object detection datasets (e.g \textit{mini}COCO is a 1\%, 5\% or 10\% subset of MS COCO). 
Several of the self-supervised object detection methods reviewed in this survey report numbers in the low-data regime; see for instance the results on \textit{mini}COCO of DETReg~\citep{detreg}, SoCo~\citep{soco}, InsLoc~\citep{insloc}, DenseCL~\citep{densecl}.
\mr{Compared to LSOD, SemiOD methods generally leverage additional unlabeled datasets. For instance,
Adaptive Class-Rebalancing~\citep{zhang2021semi} and SoftTeacher~\citep{xu2021end} report results on \textit{mini}COCO but also leverage additional unlabeled MS COCO images to improve performance on the regular MS COCO benchmark.}

\subsection{Few-shot semantic segmentation}
Few-shot object detection has many similarities to few-shot semantic segmentation as they both require us to identify the objects of interest in the images. However, semantic segmentation only considers the class of individual pixels and does not require to identify the individual objects in the image. Semantic segmentation models tend  to be simpler as they are usually based on architectures that only have a downsampling path and an upsampling path \cite{long2015fully,ronneberger2015u}, as opposed to a proposal generator and additional networks for classification and regression \cite{fastrcnn}. Further, people have explored few-shot semantic segmentation using weaker labels than the full segmentation masks. For instance, \citet{rakelly2018conditional,siam2020weakly} allow annotators to label only few pixels per object in the support or image-level labels for meta-testing. Using weakly supervised methods for few-shot object detection is is an interesting direction that is fairly unexplored.

\subsection{Zero-shot object detection}

Zero-shot object detection and instance segmentation are about learning to detect (resp. segment) novel objects based on a non-visual descriptions of them.
These descriptions could be in the form of semantic attributes, such as ``long tail'' and ``orange beak'' in the case of bird classification. In practice, the attribute vector often consists of pretrained word embeddings, since those are readily available and contain implicit world knowledge from large unlabeled datasets.
When using word embeddings, a common strategy is to modify existing object detection/instance segmentation heads by projecting object feature maps to have same dimensionality as word embeddings.
This is the case of ViLD~\citep{vild}, ZSI~\citep{zsi}, BLC~\citep{blc}, PL~\citep{pl}, DSES~\citep{dses}, who propose many tricks to improve performance, such as distilling pretrained vision-language models like CLIP~\citep{clip}, learning improved word embeddings for ``foreground'' and ``background'' for the RPN, using separate pathways for seen and unseen classes to avoid catastrophic forgetting, and explore different ways to mingle semantic and visual information. Overall, several innovations from zero-shot classification and object detection, few-shot object detection, and instance segmentation could be shared in the future.


\section{Conclusion}

We have formalized the few-shot object detection framework and reviewed the main benchmarks and evaluation metrics.
We have categorized, reviewed, and compared several few-shot and self-supervised object detection methods.
Finally, we have summarized our main takeaways, made future best practice recommendations, highlighted trends to follow, and given pointers to related tasks.


%

\section*{Acknowledgments}
This research was partially supported by the Canada CIFAR AI Chair Program, the NSERC Discovery Grant RGPIN-2017-06936, by project PID2020-120611RB-I00/AEI/10.13039/501100011033 and by Mitacs through the Mitacs Accelerate program. Simon Lacoste-Julien is a CIFAR Associate Fellow in the Learning in Machines \& Brains program.

\ifCLASSOPTIONcaptionsoff
  \newpage
\fi



\bibliographystyle{abbrvnat}
\bibliography{low-data-object-detection.bib}
%

%


\newpage

\begin{IEEEbiography}[{\includegraphics[width=1in,height=1.25in,clip,keepaspectratio]{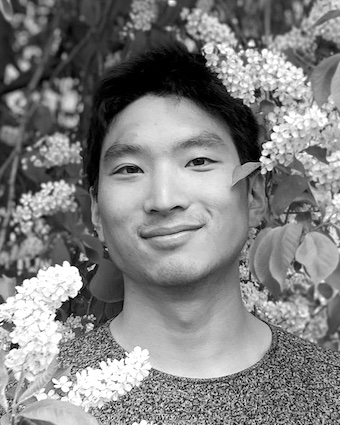}}]{Gabriel Huang}
is a PhD Candidate at Mila Québec AI Institute \& DIRO Université de Montréal advised by Simon Lacoste-Julien. He is also a Visiting Researcher at ServiceNow. His research focuses on few-shot and self-supervised learning, multimodal learning, video captioning, object detection, and language models. He strives to develop ethical AI and understand the impact of AI on society.
\end{IEEEbiography}

\begin{IEEEbiography}[{\includegraphics[width=1in,height=1.25in,clip,keepaspectratio]{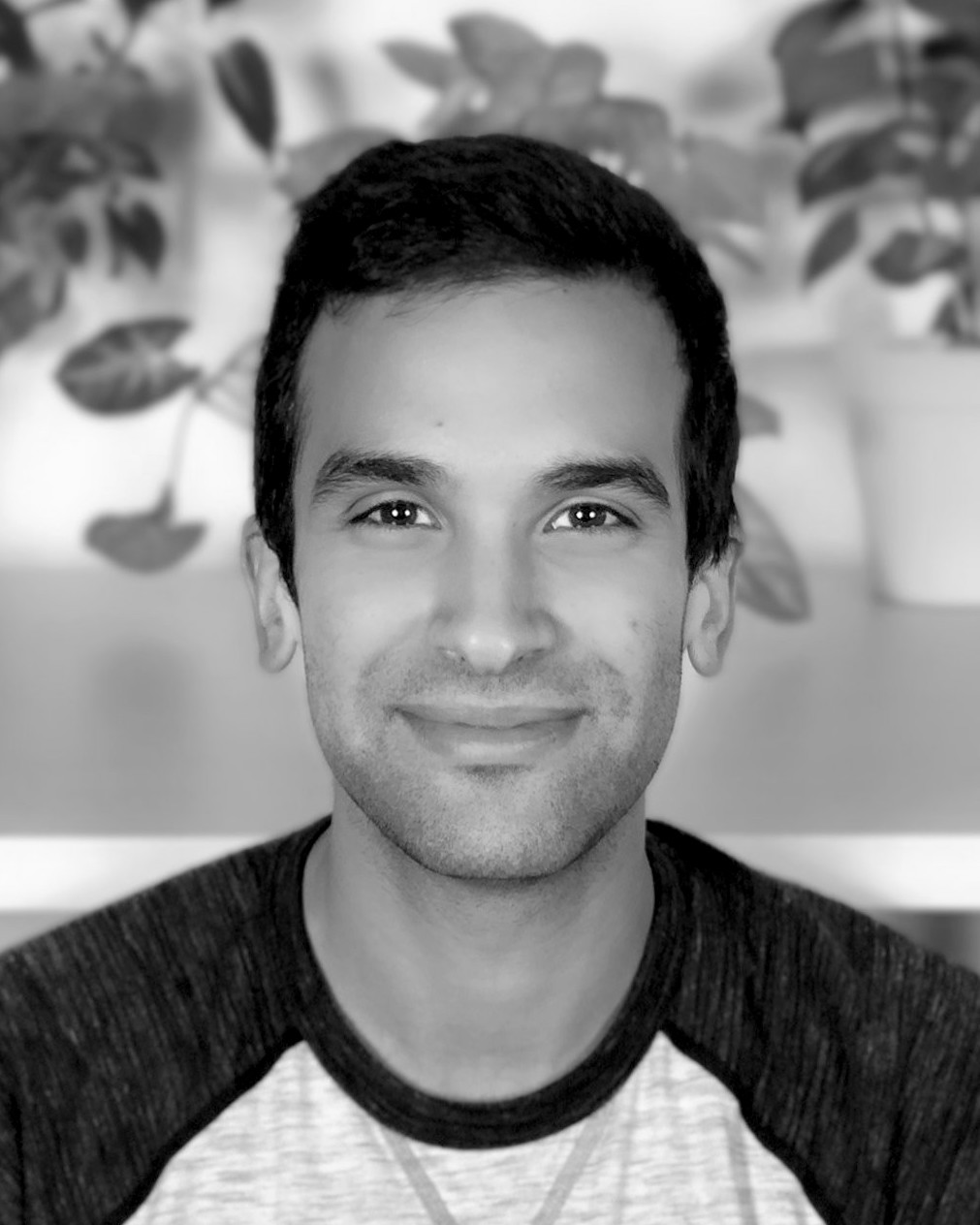}}]{\\Issam Laradji}
is a Research Scientist at ServiceNow in the low data learning lab. He completed his PostDoc at the McGill graphics lab led by Derek Nowrouzezahrai.  His main research interests are in differentiable rendering and physics, weakly supervised learning and optimization for computer vision and natural language processing applications. 
\end{IEEEbiography}

\begin{IEEEbiography}[{\includegraphics[width=1in,height=1.25in,clip,keepaspectratio]{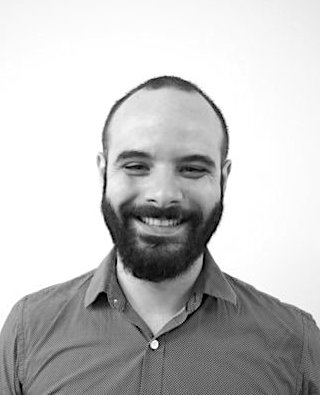}}]{\\David Vazquez}
is a Research Lead at ServiceNow working on machine learning methods that can efficiently train with few supervision applied to computer vision, and natural language processing tasks. He finished a postdoc at Montreal Institute of Learning Algorithms (Mila) and a PhD at Universitat Autonoma de Barcelona (UAB).
\end{IEEEbiography}

\begin{IEEEbiography}[{\includegraphics[width=1in,height=1.25in,clip,keepaspectratio]{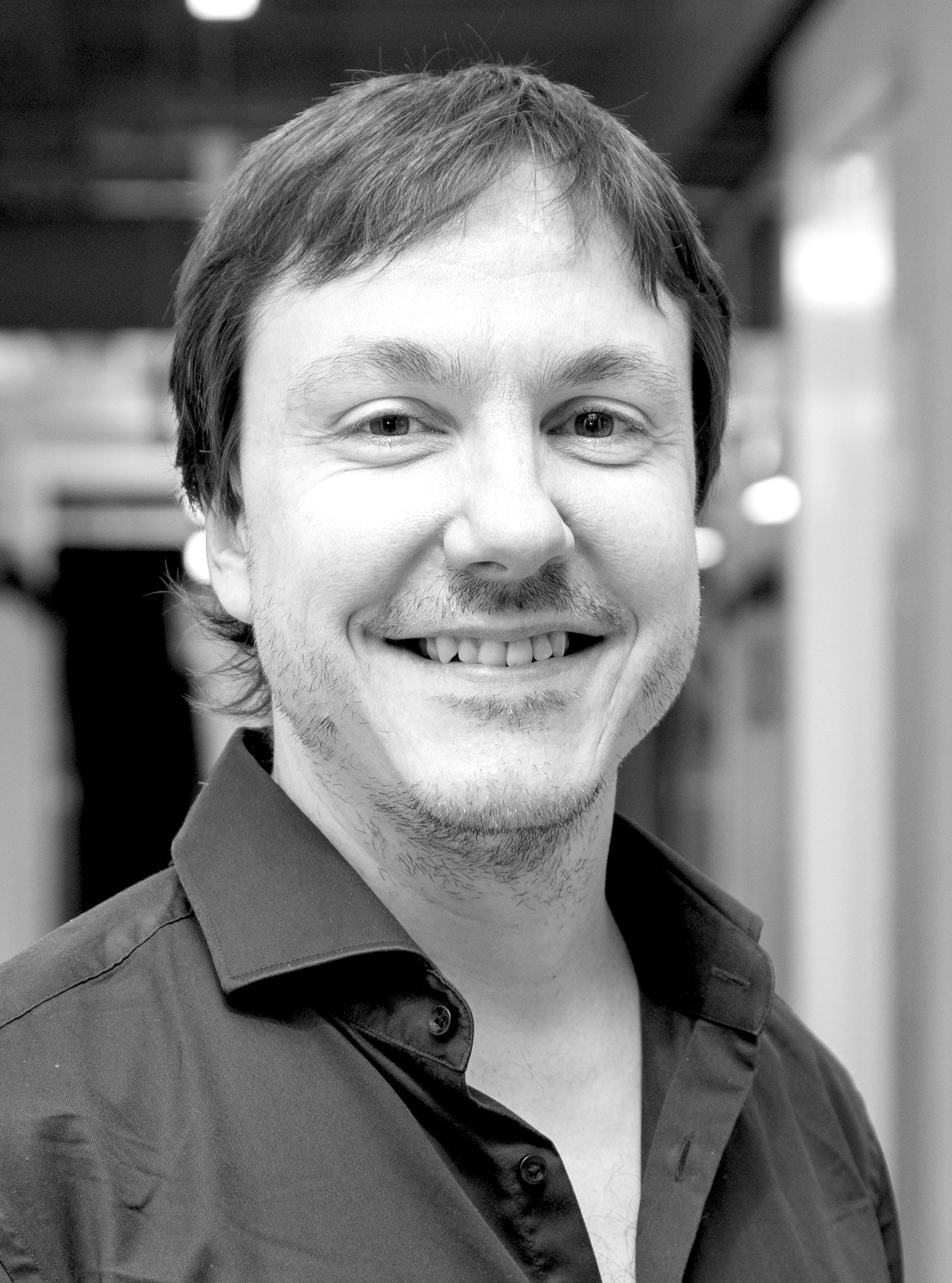}}]{Simon Lacoste-Julien}
is the associate director of the CIFAR Learning in Machines \& Brains program and  Canada CIFAR AI Chair at Mila. He is an associate professor at the Department of Computer Science and Operations Research (DIRO) at Université de Montréal and is the part-time VP Lab Director at the Samsung SAIT AI Lab in Montreal. Simon Lacoste-Julien’s research focuses on machine learning, i.e., how to program a computer to learn from data and solve useful tasks.
\end{IEEEbiography}

\begin{IEEEbiography}[{\includegraphics[width=1in,height=1.25in,clip,keepaspectratio]{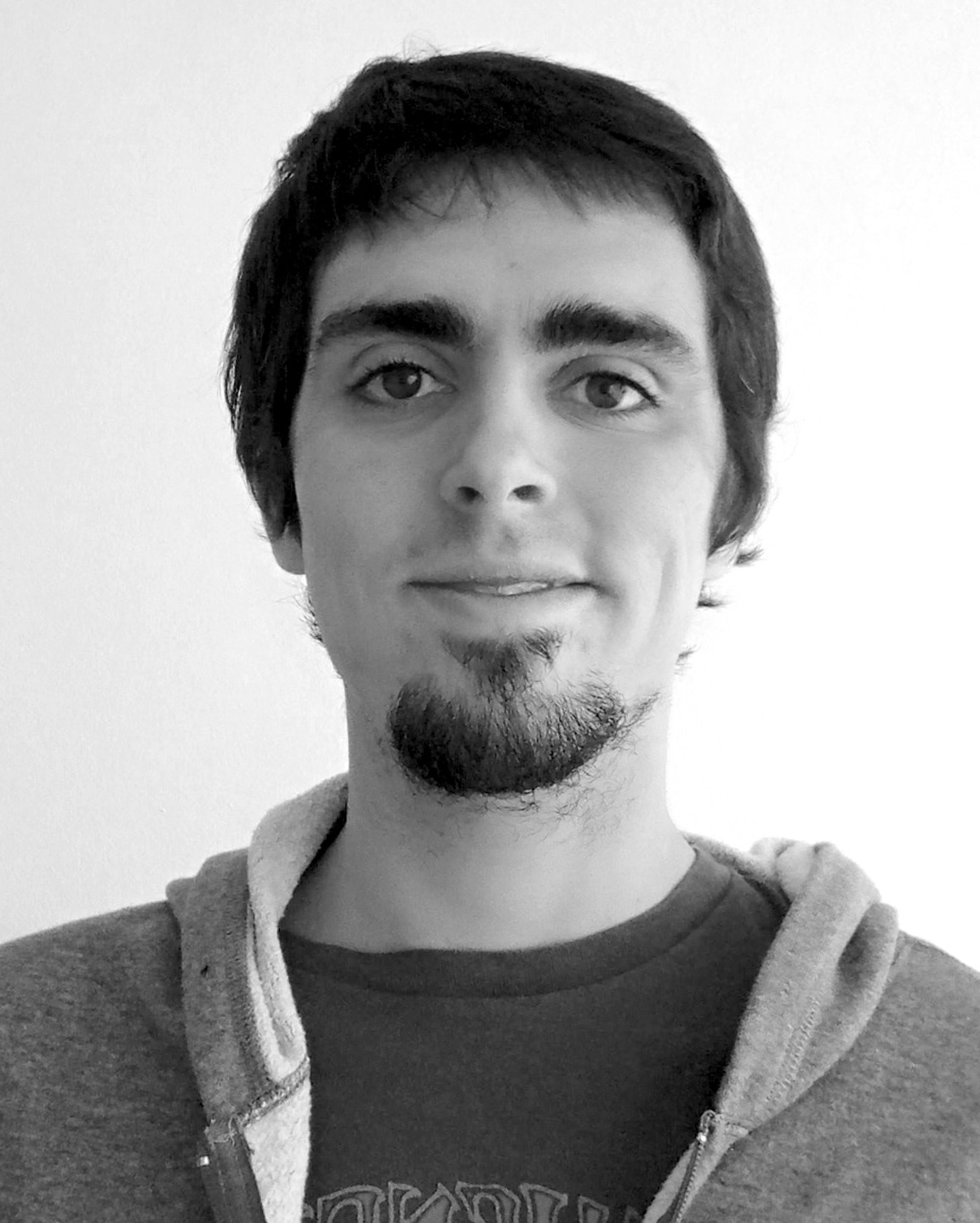}}]{\\\\\\Pau Rodriguez} is a Research Scientist at ServiceNow, adjunct professor at UAB and ELLIS member. He obtained a PhD in Computer Vision and Artificial Intelligence at UAB.
\end{IEEEbiography}







\end{document}